\UseRawInputEncoding 

\documentclass[11pt]{article}
\usepackage[letterpaper, left=1in, top=1in, right=1in, bottom=1in, verbose, ignoremp]{geometry}

\usepackage{url}\RequirePackage[colorlinks,citecolor=blue, linkcolor=blue,urlcolor = blue]{hyperref}
\usepackage{latexsym,amssymb,amsmath,amsfonts,graphicx,color,fancyvrb,amsthm,enumerate,subcaption,mathrsfs}
\usepackage[authoryear,round]{natbib}
\usepackage[dvipsnames]{xcolor}
\usepackage{xy}\xyoption{all} \xyoption{poly} \xyoption{knot}
\usepackage{float}
\usepackage{bm}
\usepackage{bbm}
\usepackage{multicol,multirow}
\usepackage{array}
\usepackage{relsize}
\usepackage{chngcntr}
\usepackage{etoolbox}
\usepackage{caption}
\usepackage{makecell}
\usepackage{tikz}
\usepackage{placeins}
\usepackage{amsthm}
\usepackage{mdframed}
\usetikzlibrary{decorations.pathreplacing,calc}
\usetikzlibrary{shapes,backgrounds}
\usetikzlibrary{patterns}
\usetikzlibrary{cd}
\usepackage{amsopn}
\usepackage{calc}
\usepackage{algorithm}
\usepackage{afterpage}
\usepackage[noend]{algpseudocode}
\usepackage{hyperref}
\usepackage{mathtools}

\usepackage{tikz-cd} 
\usepackage{booktabs}
\usepackage{amscd} 
\usepackage{comment}
\usepackage{enumitem}
\usepackage[capitalize,nameinlink,compress]{cleveref}
\crefname{figure}{Figure}{Figures} 
\crefname{equation}{}{} 
\crefname{assumption}{Assumption}{Assumptions}
\crefname{subsection}{Subsection}{Subsections}

\usepackage{graphicx} 
\usetikzlibrary{decorations.pathreplacing,backgrounds}
\usepackage{wrapfig}
\usepackage{pgf}
\usepackage{pgfplots}
\usetikzlibrary{pgfplots.external}
\usetikzlibrary{arrows}
\usetikzlibrary{arrows.meta}
\usepackage{nicematrix}
\pgfplotsset{compat=1.18}

\usepackage{notation}

\usepackage[title]{appendix}

\usetikzlibrary{calc, fit, shapes, positioning, backgrounds}

\newcounter{cdrow}

\newtheorem{theorem}{Theorem}[subsection]
\newtheorem*{theorem*}{Theorem}
\newtheorem{cor}[theorem]{Corollary}
\newtheorem{lemma}[theorem]{Lemma}
\newtheorem{prop}[theorem]{Proposition}

\newtheorem*{claim*}{Claim}
\newtheorem*{claimproof*}{Proof of claim}

\newtheorem{definition}[theorem]{Definition}

\theoremstyle{construction}

\newtheorem*{construction*}{Construction}

\theoremstyle{remark}
\newtheorem{remark}[theorem]{Remark}

\theoremstyle{example}
\newtheorem{example}[theorem]{Example}
\newtheorem*{example*}{Example}

\def\diag{{\rm diag}}
\def\log{{\rm log}}

\makeatletter
\newcommand*{\op}{%
  \DOTSB
  \mathop{\vphantom{\bigoplus}\mathpalette\matt@op\relax}%
  \slimits@
}
\newcommand\matt@op[2]{%
  \vcenter{\m@th\hbox{\resizebox{\widthof{$#1\bigoplus$}}{!}{$\boxplus$}}}%
}
\makeatother

\DeclareMathOperator{\im}{im}

\newcommand{\R}{\mathbb{R}}

\definecolor{skyblue}{RGB}{0, 180, 250}

\newcommand{\Wpre}{W_\text{pre}}

\makeatletter
\def\@biblabel#1{}
\makeatother

\makeatletter
\patchcmd{\NAT@citex}
  {\@citea\NAT@hyper@{%
     \NAT@nmfmt{\NAT@nm}%
     \hyper@natlinkbreak{\NAT@aysep\NAT@spacechar}{\@citeb\@extra@b@citeb}%
     \NAT@date}}
  {\@citea\NAT@nmfmt{\NAT@nm}%
   \NAT@aysep\NAT@spacechar\NAT@hyper@{\NAT@date}}{}{}

\patchcmd{\NAT@citex}
  {\@citea\NAT@hyper@{%
     \NAT@nmfmt{\NAT@nm}%
     \hyper@natlinkbreak{\NAT@spacechar\NAT@@open\if*#1*\else#1\NAT@spacechar\fi}%
       {\@citeb\@extra@b@citeb}%
     \NAT@date}}
  {\@citea\NAT@nmfmt{\NAT@nm}%
   \NAT@spacechar\NAT@@open\if*#1*\else#1\NAT@spacechar\fi\NAT@hyper@{\NAT@date}}
  {}{}

\makeatother

\begin{document}

\def\spacingset#1{\renewcommand{\baselinestretch}%
{#1}\small\normalsize} \spacingset{1}

\begin{flushleft}
{\Large{\textbf{Singular Geometry of Low-Rank Adaptation}}}
\newline
\\
Kamillo Ferry$^{1}$, Anthea Monod$^{1}$, and  Keisuke Yano$^{2}$
\\
 \bigskip
 \bf{1} Department of Mathematics, Imperial College London, UK
 \\
 \bf{2} Institute of Statistical Mathematics, Japan
 \\
 \bigskip
 $\dagger$ Corresponding e-mail: k.ferry@imperial.ac.uk
\end{flushleft}


\section*{Abstract}

Low-rank adaptation (LoRA, \citet{hu2021lora}) is now widely accepted as the standard approach to parameter-efficient fine-tuning of large neural networks.  Its parameter efficiency arises from a low-rank factorization that is inherently non-identifiable, making LoRA a singular statistical model where parameter dimension alone cannot describe learning complexity.  Singular learning theory overcomes the unsuitability of parameter dimension for capturing learning complexity by instead studying the real log canonical threshold (RLCT)---a geometric quantity which controls the leading asymptotic behavior of Bayesian generalization and model evidence.  We study the RLCT for LoRA and, in particular, how it depends on both the low-rank parameterization and the pretrained network in which the adaptation is performed. Under a local linearization of the pretrained network, we show that LoRA induces a low-rank trace regression model whose Kullback--Leibler geometry is controlled by a forward-backward covariance.  This covariance is constructed from the input activations and backpropagated sensitivities at the adapted layer and determines which low-rank perturbations of the pretrained model are statistically observable.  When it is nondegenerate along the relevant low-rank directions, we recover the classical learning complexity of reduced-rank regression.  When its kernel meets these directions, however, additional statistically indistinguishable parameters arise and the singular geometry of the model changes.  We characterize this geometry both locally, using a decomposition into observable tangent directions and a residual singularity, and globally, using affine linear sections of determinantal varieties.  In particular, the global learning complexity is not necessarily determined by the geometry at a ground-truth parameter; it turns out that singularities on a different rank stratum can dominate the asymptotic behavior.  To support the numerical study of this geometry, we introduce a curvature-corrected multi-temperature estimator of the RLCT that accounts for finite-sample variation across inverse temperatures.  Numerical experiments on multilayer perceptron and transformer architectures reveal strongly layer-dependent forward-backward geometry and show that ambient covariance degeneracy does not imply degeneracy along the low-rank directions relevant to adaptation.

\paragraph{Keywords:} Low-rank adaptation; singular learning theory; real log canonical threshold; low-rank matrix factorization; determinantal varieties.


\section{Introduction}
\label{sec:intro}
Low-rank adaptation (LoRA, \citet{hu2021lora}) has become the standard method for fine-tuning large neural networks due to its parameter efficiency.  Instead of updating pre-trained weight matrix $\Wpre \in \R^{m \times d}$ directly, LoRA freezes $\Wpre$ and learns the following low-rank perturbation:
$$
W = \Wpre + BA, \qquad B \in \R^{m\times r},\,  A \in \R^{r\times d}.
$$
When $r \ll \min\{m,d\}$, the number of trainable parameters reduces significantly, however, a fundamental problem of identifiability also arises, since many pairs of matrices $A,B$ are able to represent the same update $BA$.  

Approaching the problem of non-identifiability from a geometric perspective is useful in the sense that model parameters representing the same distribution may form a smooth manifold of constant dimension, so then the redundant directions can be removed locally. LoRA, however, turns out to be a \emph{singular} statistical model.  It turns out that the geometry of the factorized parameterization of LoRA changes with the rank of $BA$, which results in a stratified parameter space in which different rank strata meet along singular loci.  

Singular learning theory (SLT, \citet{Watanabe:2009,watanabe2018mathematical}) is a methodology adapted for such statistical models where complex local geometry affects the asymptotic learning behavior.  In regular models, local asymptotic behavior is controlled by a nondegenerate quadratic approximation to the Kullback--Leibler (KL) divergence and model complexity is captured simply by parameter dimension.  In singular models, this quadratic approximation breaks down because the set of parameters representing the same distribution may have positive dimension, intersect itself, or change dimension across different strata.  The asymptotics are then controlled by the local geometry of the zero set of the KL divergence and not just by the number of identifiable parameters.  SLT studies this geometry and measures it using the \emph{real log canonical threshold} (RLCT), which replaces half the parameter dimension in the leading asymptotics of Bayesian generalization and marginal likelihood.  Computing the RLCT is therefore fundamentally a geometric problem that is closely connected to the study of singular algebraic and analytic varieties.

LoRA falls within this setting of singular statistical models since its factorized parametrization has a rich singular structure.  The learning complexity of LoRA thus cannot be characterized by its nominal parameter dimension $r(m+d)$.  In addition to the low-rank geometric characterization of LoRA, the adapter is learned inside a pretrained network and the surrounding network determines which perturbations of the adapted weight matrix are distinguishable from the data.  The learning complexity of LoRA is thus made up of interactions between the intrinsic geometry of the low-rank parameterization and the statistical geometry inherited from the pretrained model.  The goal of this paper is to characterize this interaction and determine how it controls the RLCT of LoRA.

\paragraph{Contributions.} We develop a SLT framework for LoRA that explicitly characterizes the interaction between the geometry of the low-rank factorization and of the pretrained network.  Under local linearization, we identify a \emph{forward-backward covariance} which is constructed from layer activations and backpropagated sensitivities, and determines which directions of the LoRA update are statistically observable.  This leads to two qualitatively different regimes: Under suitable nondegeneracy, we recover the classical RLCT of reduced-rank regression.  In the degenerate case, the learning problem leads to affine linear sections of determinantal varieties; the different rank strata of these varieties compete to determine the asymptotic learning complexity.  We develop both the local and global theory of this interaction and show that the global RLCT need not be determined by the singularity at a ground-truth parameter.  For numerical RLCT estimation, we further introduce a \emph{curvature-corrected multi-temperature estimator} that accounts for finite-sample variation of the effective learning coefficient across inverse temperatures.  Numerical experiments on multilayer perceptron and transformer architectures support our theory and demonstrate a strong layer dependence for the forward-backward geometry in neural networks.

\paragraph{Structure of the Paper.} The remainder of the paper is organized as follows.  We close this introduction with an overview and discussion of related work. \Cref{sec:background} overviews the necessary background from SLT and the geometry of low-rank matrix factorizations.  \Cref{sec:lora_fb} formulates the local linearization of LoRA as a low-rank trace-regression problem, introduces the forward-backward covariance, and establishes conditions under which the classical RLCT of reduced-rank regression may be recovered.  \Cref{sec:local} studies the local singular geometry when these conditions fail and characterizes the contribution of statistically observable tangent directions, and reduces the remaining complexity to a residual singularity.  \Cref{sec:global} develops the global theory, using the gauge symmetry of the LoRA factorization to reduce the RLCT problem to affine linear sections of determinantal varieties and to compare singularities across different rank strata.  \Cref{sec:experiments} reviews numerical approaches to RLCT estimation, introduces our curvature-corrected multi-temperature estimator, and studies our theory empirically on multilayer perceptron and transformer architectures; in particular, we validate the local linearization and investigate the geometry of the forward--backward covariance in practice.  Finally, we close the paper with a discussion of our results for low-rank adaptation and present some open questions in \Cref{sec:discussion}.

\paragraph{Related Work.} Our work lies at the intersection of theoretical analyses of low-rank adaptation and singular learning theory.  We now review existing literature and theory on LoRA and the local-linearization perspective on fine-tuning, as well as related work on surgical fine-tuning and choice of layers to adapt.  We then position our work within the literature on RLCTs of singular statistical models and recent applications of SLT to neural networks.  Finally, since our empirical analysis requires numerical computation of the RLCT, we review existing approaches to RLCT estimation and their relation to the estimator introduced in this work.

\paragraph{Theory of LoRA.}
Theoretical understanding of LoRA remains limited.
\citet{zeng2024} characterized its expressive power,
and the convergence of LoRA training has been analyzed by \citet{dayi2024,kim25n,zhang2025}.
However, its generalization behavior from the SLT viewpoint has not been investigated.

\paragraph{NTK View of Fine-Tuning.}
The local linearization in \Cref{sec:lora_fb}, a key assumption of our analysis, is the approximation underlying the neural tangent kernel (NTK) view \citep[cf.][]{jacot2018}.
\citet{malladi23a} empirically showed that the NTK view describes prompt-based fine-tuning of language models well.
Under the NTK view, \citet{jang24d} proved that LoRA training has no spurious local minima.

\paragraph{Surgical Fine-Tuning.}
Fine-tuning only a subset of layers, which we focus on in this paper,
is known as \emph{surgical fine-tuning} \citep[c.f.,][]{lee2023,lodha2023,ando2026}.
How to choose which layers to adapt is still an open question;
for instance, \citet{lodha2023} rank layers by their Fisher information while \citet{ando2026} rank them by the similarity of internal representations.
Our analysis of how the RLCT depends on the choice of adapted layers
sheds light on this question from the SLT viewpoint.

\paragraph{RLCT of Singular Models.}
The RLCT or its upper bound has been computed analytically for various singular models:
mixture models \citep{yamazaki2003},
reduced rank regression \citep{Aoyagi+Watanabe:2005},
and its extension to deep linear networks \citep{aoyagi2024},
nonnegative matrix factorization \citep{hayashi2017},
linear networks with a concept bottleneck structure \citep{hayashi2025},
Gaussian latent tree models \citep{Drton2017} and factor models \citep{drton2025},
and three-layer neural networks \citep{watanabe2001,aoyagi2005,aoyagi2012,kurumadani2026}.
\citet{watanabe2026} studied the distinction between the RLCT of each layer and that of the whole network,
i.e., layer-wise versus joint learning.
Beyond model-specific derivations, \citet{sergeant2026} proposed an algorithm that computes the local RLCT exactly for any model with two parameters.

Since LoRA parameterizes the weight update as a low-rank product,
it is natural to expect its RLCT to coincide with that of reduced rank regression \citep{Aoyagi+Watanabe:2005}.
We show that this holds under non-degeneracy of the forward-backward covariance (\Cref{sec:nondegenerate}),
and that the two differ in general, both theoretically (\Cref{sec:local} and \Cref{sec:global}) and empirically (\Cref{sec:experiments}).

Beyond analytical derivations, the relevance of SLT to deep learning
has been argued by \citet{wei2022}, and the \emph{local learning coefficient} (LLC)---an estimable local version of the RLCT---has been applied to large models \citep{lau2025}
and used to track the developmental stages of transformers during training \citep{hoogland2025}.
Closely related to our work,
\citet{wang2025} refined the LLC to subsets of weights, such as individual attention heads,
to compare the complexity of different components.
We analytically show that the RLCT of the LoRA parameters in the adapted layers
depends on the frozen pretrained network through the forward-backward covariance,
and hence cannot be determined by the adapted layers alone.

\paragraph{Estimation of the RLCT.}
Since the RLCT is analytically available only for limited classes of models,
it is often estimated from posterior samples at inverse temperatures around $1/\log n$,
based on the asymptotic expansion of the tempered posterior mean of the loss \citep{Watanabe:2013}.
Single-temperature estimators use the posterior at $\beta^{\star}=1/\log n$ only:
the WBIC-type estimators measure the gap between the posterior mean of the loss and either its value at a local minimizer \citep{lau2025} or the empirical loss of the Bayes predictive distribution \citep{takio2025}.
The heat-capacity estimator \citep{imai2019estimating,tokuda2022}
uses the posterior variance of the loss.

Multi-temperature estimators instead exploit how the tempered posterior mean of the loss varies with the inverse temperature:
\citet{Watanabe:2013} proposed a two-temperature estimator,
and \citet{wei2022} regressed the tempered posterior mean of the loss on $\beta^{-1}$ over a set of inverse temperatures.
For finite $n$, however, the tempered posterior mean of the loss need not be linear in $\beta^{-1}$,
especially for deep models, which makes these estimators sensitive to the choice of temperatures
\citep[cf.][for the two-temperature estimator]{takio2025}.
Our curvature-corrected multi-temperature estimator (\Cref{sec:experiments}) mitigates this sensitivity
by a local quadratic fit in $\beta^{-1}$.

\section{Preliminaries on Singular Learning Theory}
\label{sec:background}
In this section, we briefly overview and summarize the tools from SLT used throughout the paper.  We first recall the real log canonical threshold (RLCT) and the basic properties that allow us to study it locally and using analytic ideals.  Our main technical tool is a local splitting principle: after suitable analytic changes of coordinates, the regular directions of a statistical model can be separated from the singular part.  The regular directions contribute to the RLCT in the usual dimension-counting manner, while the remaining contribution is encoded by the singularity of a lower-dimensional residual model.

\subsection{Real Log Canonical Thresholds}

Let $q(x)$ denote the true data-generating distribution and $p(x \mid w)$, $w \in \Omega$ denote a statistical model.  A natural measure of how well the parameter $w$ represents the truth is the Kullback--Leibler (KL) divergence
\[
K(w)
=
\int q(x) \log \frac{q(x)}{p(x \mid w)} \intDifferential{x}.
\]
Since $K(w) \geq 0$, its zero set
\[
\{w \in \Omega : K(w) = 0\}
\]
consists of the parameters for which the model represents the true
distribution.  In a regular identifiable model, locally, this zero set is an
isolated point and $K$ has a nondegenerate quadratic approximation
around it.  In a singular model, however, the zero set may have positive dimension or singularities; $K$ may also vanish at different rates in different directions.

This local behavior of $K$ determines how probability mass accumulates around parameters representing the truth.  In Bayesian learning, the marginal likelihood is obtained by integrating the likelihood over parameter space and as the sample size grows, its leading asymptotic contribution comes from neighborhoods where $K(w)$ is small.  The RLCT measures the rate of this concentration.  In regular models, this rate is given simply by half the parameter dimension; in
singular models, it depends on the geometry of $K$ near its zero set.

This behavior can be studied using the \emph{zeta function} associated with $K$.  Let $\Omega \subset \RR^n$ be a compact semianalytic set and let $\psi \colon \Omega \to \RR$ be nearly analytic.  For an
analytic function $K \geq 0$, $K \not\equiv 0$, consider the zeta function
\[
  \zeta(z) = \int_\Omega K(w)^{-z}\abs{\psi(w)} \intDifferential{w}.
\]
Here \(\psi\) is nearly analytic, i.e., it is a product of a real-analytic function and a positive smooth function \citep{MR3614481}.
Throughout our work, we assume that the original prior \(\psi\) is real-analytic and strictly positive in a neighborhood of the zero set of 
\(K\). Then coordinate changes multiply \(\psi\) by a Jacobian determinant which yields a nearly analytic function.

The function $\zeta$ admits a meromorphic continuation to the whole complex plane.  The poles of this continuation are positive rational numbers \citep{MR256156} and encode the rate  at which $K$ vanishes near its zero set; the smallest pole defines the RLCT.

\begin{definition}
  Let \(\lambda\) be the smallest pole for the continuation of \(\zeta(z)\) and \(\multiplicity\) be its multiplicity
  as a pole. Then the pair \((\lambda,\multiplicity)\) is called the \emph{real log canonical threshold}
  of \(K\) with respect to \(\psi\) and \(\Omega\) and is denoted by \(\rlct_\Omega(K)\).
\end{definition}

In our settings of interest, the KL divergence will often be equivalent to a sum of squares of analytic functions, up to positive multiplicative
constants:
\[
K(w) \asymp \sum_{i=1}^{\nu} f_i(w)^2.
\]
Its zero set is therefore the common zero set of the functions $f_1,\ldots,f_\nu$.  Rather than taking a particular choice of these functions to be part of the statistical model, it will be useful to work with
the analytic ideal that they generate:
\[
I = \langle f_1,\ldots,f_\nu\rangle.
\]
Working with such an ideal is convenient because the same ideal, and therefore the same local geometry, may be described by different sets of generators.  In this case, it will be useful to extend the definition of the RLCT from a single analytic function to an analytic ideal.

\begin{definition}
\label{def:rlct_ideal}
  Let \(I=\langle\,f_1,\dots,f_\nu\,\rangle\) be an ideal of the ring \(\mathcal{A}_\Omega\) of real-analytic functions on \(\Omega\). Then the \emph{real log canonical threshold of \(I\)}
  is the real log canonical threshold of the zeta function \[
    \zeta(z) = \int_\Omega \left(
    \sum_{i=1}^\nu f_i^2
  \right)^{-z}
  \abs{\psi(w)} \intDifferential{w}
  \] and is denoted by \(\rlct(f_1,\dots,f_\nu)\).
\end{definition}

For Definition \ref{def:rlct_ideal} to depend only on the ideal $I$, rather than on the particular functions that can be chosen to generate it, the following standard invariance property is required.

\begin{prop}[{\citealt[Prop.~3.6]{MR3614481}}]\label{prop:ideal-independence}
    For two generating sets 
    \(I = \langle\, f_1,\dots,f_\nu \,\rangle = \langle\, g_1,\dots,g_{\nu'} \,\rangle\)
    of an ideal \(I\), the pairs \(\rlct(f_1,\dots,f_\nu)\) and \(\rlct(g_1,\dots,g_{\nu'})\)
    agree. Thus, the RLCT \(\rlct(I)\) for an ideal is well defined and 
    independent of the chosen generating set of \(I\). 
\end{prop}

This result allows us to replace one collection of analytic equations by another whenever they generate the same ideal without changing the RLCT.

We will also often separate collections of generators that depend on disjoint sets of variables.  In this setting, their RLCTs combine in the following manner according to the product rule.

\begin{lemma}[{\citealt[Prop.~3.7]{MR3614481}}]\label{lem:sum-product-rule}
    Let \(f_1,\dots,f_\nu\) be analytic functions on a compact semianalytic set \(X\), and \(g_1,\dots,g_{\nu'}\) 
    analytic functions on a compact semianalytic set \(Y\). Let \(\psi_X\) and \(\psi_Y\) be nearly analytic with
    \(\rlct(f_1,\dots,f_\nu)=(\lambda_X,\multiplicity_X)\) on \(X\) and \(\rlct(g_1,\dots,g_{\nu'})=(\lambda_Y,\multiplicity_Y)\) on \(Y\).
    Then, regarded as functions on \(X\times Y\) with respect to \(\psi_X\psi_Y\), \[
        \rlct(f_1,\dots,f_\nu,g_1,\dots,g_{\nu'})=(\lambda_X+\lambda_Y,\ \multiplicity_X+\multiplicity_Y-1),
    \] and \(\rlct(\{f_ig_j\}_{i,j})\) equals \((\lambda_X,\multiplicity_X)\) if \(\lambda_X<\lambda_Y\), equals \((\lambda_Y,\multiplicity_Y)\) if 
    \(\lambda_X>\lambda_Y\), and equals \((\lambda_X,\multiplicity_X+\multiplicity_Y)\) if \(\lambda_X=\lambda_Y\).
\end{lemma}

The zero set of a singular model may contain  parameters with different local geometry, so understanding the model globally requires understanding its singularities point by point.  For a point $w_0 \in \Omega$, the \emph{local RLCT}
\[
\rlct_{w_0}(I)
=
(\lambda_{w_0},\multiplicity_{w_0})
\]
is obtained by restricting the integral to a sufficiently small neighborhood of $w_0$.  The local RLCT therefore measures the singularity of the statistical model specifically near $w_0$ and the contribution to the asymptotic learning behavior of the model.

We order RLCT pairs by
\[
(\lambda,\multiplicity) < (\lambda',\multiplicity')
\]
if either $\lambda<\lambda'$, or $\lambda=\lambda'$ and $\multiplicity>\multiplicity'$.  With this convention, the global RLCT is the minimum of the local thresholds,
\[
\rlct_{\Omega}(I)
=
\min_{w\in V(I)}
\rlct_{w}(I),
\]
where
\(V(I) = \{w\in\Omega:f(w)=0\text{ for every }f\in I\}\) is the zero set of the ideal. It is important to note the multiplicity convention in the ordering, because when two local thresholds have the same value of $\lambda$, the larger multiplicity $\multiplicity$ is dominant in the asymptotic contribution.

The statistical significance of the global RLCT can be seen when comparing with the classical Bayesian information criterion (BIC).  For a regular statistical model with $k$ parameters, the BIC 
\citep{Schwarz:1978} is
\[
\mathrm{BIC}
=
-\sum_{i=1}^n \log p(X_i \mid \widehat{w})
+
\frac{k}{2}\log n,
\]
where $\widehat{w}$ is the maximum likelihood estimator.  Up to a bounded term, this agrees with the asymptotic expansion of the \emph{free energy} $F_n$, i.e., the negative logarithm of the marginal
likelihood:
\[
F_n
=
nS_n
+
\frac{k}{2}\log n
+
O_P(1),
\]
where $S_n$ is the empirical entropy of the true distribution.  In a regular model, model complexity therefore enters the Bayesian asymptotics through the parameter count $k$.

For singular models, such a dimension-counting interpretation does not hold.  The set of parameters representing the true distribution is not necessarily given by only an isolated point and the Fisher information may be degenerate.  In this case, the appropriate correction to the free energy is determined by the RLCT.

\begin{theorem}[{\citealt[Theorem~2]{Watanabe:2013}}]
\label{thm:free-energy}
Let $K$ be the KL divergence from the true distribution
to the model, with prior density $\psi$ on $\Omega$, and suppose that
the fundamental conditions of \citet{Watanabe:2013} hold. If
\[
\rlct_\Omega(K)=(\lambda,\multiplicity),
\]
then the free energy satisfies
\[
F_n
=
nS_n
+
\lambda\log n
-
(\multiplicity-1)\log\log n
+
O_P(1).
\]
\end{theorem}

For a regular model, $\lambda=k/2$ and $\multiplicity=1$ \citep[Lemma~3]{Watanabe:2013}, so the theorem recovers the classical BIC asymptotics.  In a singular model, the RLCT replaces parameter dimension as the appropriate measure of model complexity.

The RLCT also controls predictive performance \citep{Watanabe:2009}: Let $G_n$ denote the
Bayes generalization error, i.e., the KL divergence from
the true distribution to the posterior predictive distribution.  Then
\[
\mathbb{E}[G_n]
=
\frac{\lambda}{n}
+
o\left(\frac{1}{n}\right).
\]
Thus, the geometry encoded by the RLCT has a
direct statistical interpretation in terms of both model evidence and
predictive generalization.

\subsection{Contact Equivalence and Local Splitting}
\label{sec:contact_splitting}

The local RLCT is invariant under suitable analytic changes of coordinates, which allows us to simplify the local geometry of a singular model before computing its threshold.  In particular, suppose that near a point $w_0$, the KL divergence is equivalent to
\[
K(w) \asymp \norm{F(w)}^2
\]
for a real-analytic map
\[
F \colon (\RR^N,w_0) \longrightarrow (\RR^D,0).
\]
If the differential \(\differential F_{w_0}\) has rank $s$, then $s$ directions are already visible to the first order and behave locally like regular parameters.   The remaining directions contain the essentially singular part of the model.  Our aim is to separate these two contributions: after suitable analytic changes of coordinates, the regular directions will contribute $s/2$ to the RLCT, and then what remains to be analyzed will be the singularity of a lower-dimensional residual model.

In order to achieve our aim and make this reduction precise, we will use the notion of \emph{contact equivalence}.  Contact equivalence allows for both an analytic change of coordinates in the parameter space as well as an invertible analytic change of generators of the defining ideal.  Both operations preserve the local RLCT.

\begin{definition}
\label{def:contact-equivalence}
Two germs \(f,g\colon(\RR^n,x_0)\to(\RR^p,0)\) of real-analytic maps are \emph{contact equivalent}
if there exist a real-analytic diffeomorphism germ \(\Xi\colon(\RR^n,x_0)\to(\RR^n,x_0)\)
and a real-analytic germ \(P\colon(\RR^n,x_0)\to\GenLinGrp{p}\) such that
\(g=P\cdot(f\circ\Xi)\).
\end{definition}

Thus, contact equivalence allows us to reparametrize the domain and replace the components of the defining map by invertible analytic combinations of one another.  Geometrically, these operations change
the local representation of the singularity without changing the singularity relevant to the RLCT.

\begin{lemma}\label{lem:contact-invariance}
    If two germs \(f,g\colon(\RR^n,x_0)\to(\RR^p,0)\) are contact equivalent, then 
    \[\rlct_{x_0}(g_1,\dots,g_p)=\rlct_{x_0}(f_1,\dots,f_p).\]
\end{lemma}

\begin{proof}
        Let \(g=P\cdot(f\circ\Xi)\). On a compact neighborhood of \(x_0\) on which \(P\) and \(P^{-1}\) are analytic, the components 
        of \(g\) and of \(f\circ\Xi\) are analytic combinations of each other, so they generate the same ideal. Hence 
        \(\rlct_{x_0}(g_1,\dots,g_p)=\rlct_{x_0}(f_1\circ\Xi,\dots,f_p\circ\Xi)\) by Proposition \ref{prop:ideal-independence}, and this equals 
        \(\rlct_{x_0}(f_1,\dots,f_p)\) by the invariance of local thresholds under real-analytic changes of variables 
        \cite[Prop.~3.8 and Lemma~2.4]{MR3614481}.
\end{proof}

Here, contact equivalence is useful because it allows us to separate directions that are already regular to first order from those responsible for the singularity.  If \(\differential F_{w_0}\) has rank $s$, then locally there are $s$ independent directions where $F$ varies linearly.  After appropriate analytic changes of coordinates and generators, these directions can be isolated as coordinate functions; meanwhile, all remaining nonlinear behavior is confined to a residual map whose differential vanishes at the origin.  The RLCT can then be decomposed into the regular contribution from these
$s$ directions and the contribution coming from the residual singularity.  The following splitting lemma formalizes this decomposition.

\begin{lemma}\label{lem:splitting}
    Let \(F\colon(\RR^N,x_0)\to(\RR^D,0)\) be a germ of a real-analytic map and let \(s=\rank\differential F_{x_0}\).
    Then \(F\) is contact equivalent to the germ
    \((x,y)\mapsto (x,\chi(y))\) at \((0,0)\in\RR^s\times\RR^{N-s}\) for some 
    real-analytic \(\chi\) with values in \(\RR^{D-s}\), \(\chi(0) = 0\) and \(\differential \chi(0) = 0\).
    In particular, \(\norm{F}^2\asymp\norm{x}^2+\norm{\chi(y)}^2\) in suitable real-analytic coordinates \((x,y)\) centered at \(x_0\).
\end{lemma}

    \begin{proof}
        If \(s = 0\), there are no regular coordinates and we can take \(\chi = F\); here \(\chi(0)=0\) since
        \(F(x_0)=0\) and \(\differential \chi(0)=0\) since \(\differential F_{x_0}=0\). Thus, assume \(s \geq 1\).
        In this case, there is a non-vanishing \(s\times s\)-minor
        of \(\differential F_{x_0}\). 

        Relabel the source coordinates as \((u_1,u_2)\) and the target components so
        that \(F=(F_{\mathrm{reg}},F_{\mathrm{res}})\)
        with \(\det(\partial F_{\mathrm{reg}}/\partial u_1)(x_0)\neq0\).
        Set \(\Xi(u_1,u_2)=(F_{\mathrm{reg}}(u_1,u_2),u_2)\). In particular, the Jacobian of this coordinate change is non-vanishing. Thus, by the real-analytic inverse function theorem it is a local real-analytic
        diffeomorphism near \(x_0\). Write \(x = F_{\mathrm{reg}}\) and \(y = u_2\).
        By the resulting local coordinate change, we obtain \[
            F = (x,\chi(x,y)),
        \] where  \(\chi(0,0) = 0\) since \(F(x_0)=0\).

        By the analytic version of Hadamard's lemma, \(\chi(x,y) - \chi(0,y) = Q(x,y)x\)
        where \(Q\) is an analytic matrix-valued function. Set 
        \(P(x,y)=\begin{psmallmatrix}I&0\\-Q(x,y)&I\end{psmallmatrix}\), which is real-analytic and invertible. Then \[
            P(x,y)\cdot(x,\chi(x,y)) = (x,\chi(x,y)-Q(x,y)x) = (x,\chi(0,y)).
        \] 
        
        Set \(\chi(y) = \chi(0,y)\). Then, \(\chi(0) = \chi(0,0) = 0\). Since \(\Xi\) is a diffeomorphism and \(F(x_0)=0\), the differential of \(P\cdot(F\circ\Xi^{-1})\) at \((0,0)\) is
        \(P(0,0)\cdot\differential(F\circ\Xi^{-1})_{(0,0)}\), so \((x,\chi)\) has the same rank \(s\) at \((0,0)\) as \(F\) does at \(x_0\).
        
        Row-reducing the differential of \((x,\chi)\) then forces \(\partial_y\chi(0,0) = 0\),
        meaning \(\differential \chi(0) = 0\).
        Both \(P\) and \(P^{-1}\) are real-analytic hence bounded near \((0,0)\), 
        which means \(\norm{F}^2 \asymp \norm{x}^2 + \norm{\chi(y)}^2\).
    \end{proof}

Lemma \ref{lem:splitting} is the real-analytic version of two statements from \cite{MR4321457} which are stated there for \(C^\infty\) and holomorphic germs. The first puts a germ of rank \(s\) at a point into the form \((x,y)\mapsto(x,\chi(x,y))\) \cite[Lemma~2.3 and Exercise~2.3.6]{MR4321457}.  The second says that every unfolding is contact equivalent to the constant unfolding \cite[Cor.~4.2(1)]{MR4321457}. 
    
We give the argument in the real-analytic setting, with an explicit matrix in place of  the ideal-theoretic argument of \cite[Prop.~4.5]{MR4321457}, since the explicit matrix gives the norm equivalence directly.

\begin{cor}\label{cor:splitting-rlct}
    In the setting of Lemma \ref{lem:splitting}, \(\rlct_{x_0}(F_1,\dots,F_D)=(\lambda_{x_0},\multiplicity_{x_0})\) is given by \[
        \lambda_{x_0} = \frac{s}{2} + \lambda_\chi
        \quad\text{and}\quad
        \multiplicity_{x_0} = \multiplicity_\chi,
    \] where \((\lambda_\chi,\multiplicity_\chi)=\rlct_0(\chi_1,\dots,\chi_{D-s})\), or \((\lambda_\chi,\multiplicity_\chi)=(0,1)\) if \(\chi\equiv0\).
\end{cor}

    \begin{proof}
        By Lemma \ref{lem:contact-invariance}, the local threshold of \(F\) at \(x_0\) 
        equals the local threshold of \((x,\chi(y))\) at the origin, that is, of the ideal generated by \(x_1,\dots,x_s,\chi_1,\dots,\chi_{D-s}\). 
        Since the amplitude is positive there, we may take it to be \(1\) 
        \cite[Lemma~2.4]{MR3614481}. As \(x\) and \(y\) are disjoint sets of variables, 
        Lemma \ref{lem:sum-product-rule} applies on a small product 
        neighborhood of the origin. 
        
        For each coordinate \(x_i\), the zeta function \(\int(x_i^2)^{-z}\intDifferential{x_i}\) 
        has a simple pole at \(z=\frac12\), so \(\rlct_0(x_i)=(\frac12,1)\) and \(\rlct_0(x_1,\dots,x_s)=(\frac s2,1)\). 
        This proves the claim. If \(s=0\) or \(\chi\equiv0\), one of the two groups of
        generators is empty and there is nothing to add.
    \end{proof}

\section{LoRA and Forward-Backward Geometry}
\label{sec:lora_fb}
We now adapt the singular learning framework given in \Cref{sec:background} to low-rank adaptation.  The factorization \(BA\) determines the singular geometry of the LoRA parametrization, but by itself, it does not determine the statistical geometry of the adapted model.  This is because the update is applied to a pretrained network, so its effect on the model output also depends on the activations entering the adapted layer and on how perturbations at that layer propagate through the remainder of the network.

In this section, we will study this interaction in detail.  Under a local linearization of the pretrained network, we first formulate the learning problem for a single LoRA-adapted layer as a low-rank trace-regression model.  Its KL divergence is controlled by a covariance operator constructed from the forward activations and backpropagated sensitivities of the network, which we will refer to as the \emph{forward--backward (FB) covariance}.  Its geometry determines which perturbations of the adapted weight matrix are statistically distinguishable and therefore provides an interface between the low-rank geometry of LoRA and the statistical geometry derived from the pretrained model.

We also identify a nondegenerate regime in which the FB covariance preserves all low-rank directions relevant to the adaptation problem.  In this case, the KL divergence is equivalent to the Frobenius-norm loss of reduced-rank regression which recovers the classical RLCT.  The failure of this nondegeneracy condition and the additional singular geometry that results will be further studied in \Cref{sec:local} and \Cref{sec:global}.

\subsection{Local Linearization and Low-Rank Trace Regression}

We consider LoRA applied to a single linear layer with frozen pretrained weight \(W_{\mathrm{pre}}\in\RR^{m\times d}\).
The layer is updated as \(W=W_{\mathrm{pre}}+\Delta W\) with \(\Delta W=BA\) for \(B\in\RR^{m\times r}\) and 
\(A\in\RR^{r\times d}\), where the usual scaling factor is absorbed into \(B\).
Let \(f_W(x)\) denote the output of the network on input \(x\) when the weight of this layer is replaced by \(W\); 
all other parameters remain frozen.
Let \(\forward(x)\in\RR^d\) be the input activation of the layer and let \[
    \backward(x) \coloneqq \left.\frac{\partial f_W(x)}{\partial z^\ell}\right\vert_{W=W_{\mathrm{pre}}}\in\RR^m
\] be the backpropagated sensitivity with respect to the layer output \(z^\ell=W_{\mathrm{pre}}\forward(x)\). Thus, \(\forward(x)\) describes the forward signal entering the layer and \(\backward(x)\) describes the first-order sensitivity of the network output to perturbations of the layer output.
To first order, the update \(\Delta W\) therefore changes the network output by
$$
\frobprod{
\backward(x)\forward(x)^\transpose,\Delta W
}.
$$
This motivates the following feature representation.

\begin{definition}[Forward--backward feature]
\label{def:fb_feature}
For an input \(x\), the \emph{forward--backward (FB) feature} at the adapted layer is
$$
\Phi(x)
\coloneqq
\backward(x)\forward(x)^\transpose
\in\RR^{m\times d}.
$$
\end{definition}

With this notation, the local change in network output is
$$
f_W(x)-f_{W_{\mathrm{pre}}}(x)
\approx
\frobprod{\Phi(x),\Delta W}.
$$
Fine-tuning a single layer with LoRA is thus a reduced-rank regression problem in the sense that the response depends linearly on the 
coefficient matrix \(BA\) of rank at most \(r\), through the feature matrix \(\Phi(x)\).
We make the following assumptions, which are used throughout our work.
\begin{enumerate}[label=(A\arabic*)]
    \item \textbf{The output is scalar and the linearization above is exact.} Then the model for the residual target 
        \(\widetilde y\coloneqq y-f_{W_{\mathrm{pre}}}(x)\in\RR\) is the reduced-rank regression model \[
            p(\widetilde y\mid x,A,B)=\frac1{\sqrt{2\pi}}\exp\left(-\frac12\bigl(\widetilde y-\frobprod{\Phi(x),BA}\bigr)^2\right).
        \]
    \item \textbf{The model is realizable.}  The data are generated by \(p(\,\cdot\mid x,A_0,B_0)\) for some true parameters \((A_0,B_0)\).
    \item \textbf{The FB feature has finite second moment} \(\mathbb{E}\norm{\Phi(X)}^2<\infty\) for \(X\) drawn from the input distribution.
\end{enumerate}
Let \(D\coloneqq md\), \(S\coloneqq BA-B_0A_0\) and \(\rho\coloneqq\rank B_0A_0\). Then \(S\) has rank at most \(r+\rho\).
Since both models are Gaussian with unit variance, the KL divergence 
between the true model and the model at \((A,B)\), averaged over the inputs, is
$$
    K(A,B) = \frac12\mathbb{E}\frobprod{\Phi(X),S}^2.
$$
The quadratic form appearing here is determined by the second moment of the FB feature, which then motivates the following definition.

\begin{definition}[Population forward--backward covariance]
\label{def:fb_cov}
The \emph{population forward--backward (FB) covariance} at the adapted layer is
$$
\populationFB
\coloneqq
\mathbb{E}
\left[
\vector{\Phi(X)}
\vector{\Phi(X)}^\transpose
\right]
\in\RR^{D\times D},
$$
where the expectation is taken with respect to the input distribution.
\end{definition}

We make the important remark here that the population FB covariance $C$ is \textit{not} the Fisher information matrix.  The Fisher information matrix $M_{\mathrm{FI}} \in \mathbb{R}^{(rd+mr)\times(rd+mr)}$
for $(\operatorname{vec}A, \operatorname{vec}B)$ under the model (A1) is given by
\[
M_{\mathrm{FI}} =
\begin{pmatrix}
(I_{d} \otimes B^{\top})\, C\, (I_{d} \otimes B) & (I_{d} \otimes B^{\top})\, C\, (A^{\top} \otimes I_{m}) \\
(A \otimes I_{m})\, C\, (I_{d} \otimes B) & (A \otimes I_{m})\, C\, (A^{\top} \otimes I_{m})
\end{pmatrix}.
\]
Rather than working with $M_{\mathrm{FI}}$, in which $C$ and $(A,B)$ are entangled,
we treat $C$ and $(A,B)$ separately, which leads to the following finer analysis.


In terms of \(\populationFB\), the KL divergence is
$$
    K(A,B) = \frac12\vector{S}^\transpose\populationFB\vector{S}.
$$
The population FB covariance is a second-moment matrix; in particular, it is real, symmetric and positive semi-definite. 

\begin{prop}
\label{prop:rlct_aw}
Suppose $\populationFB$ is positive definite.  Then
$$
K(A,B)\asymp\frobnorm{BA-B_0A_0}^2
$$
and therefore the RLCT of the locally linearized LoRA model coincides with the RLCT of the corresponding reduced-rank regression model.
That is, the RLCT of the locally linearized LoRA model with \(\rho = \rank{B_0A_0}\) is 
\begin{equation}\label{eq:aoyagi-watanabe}
\lambda = \frac12\min\left\{\,
    (m+d)\rho-\rho^{2}+s(m-\rho)+(d-\rho-s)(r-\rho-s)
    \mid
    0\leq s\leq \min\{d,r\}-\rho 
\,\right\}
\end{equation}
and 
\[
  \multiplicity =
  \begin{cases}
    2 & \text{if } m+d+r+\rho \text{ is odd and } \rho\le\min\{d+r-m,\;m+r-d,\;m+d-r\},\\
    1 & \text{otherwise.}
  \end{cases}
\]
\begin{proof}
    Comparability between \(K(A,B)\) and \(\frobnorm{BA-B_0A_0}^2\) is given by the statement of \cite[Lemma 1]{Aoyagi+Watanabe:2005}.
    The main theorem of \cite{Aoyagi+Watanabe:2005} then gives the expression for \(\lambda\).
    Since the expression in \eqref{eq:aoyagi-watanabe} is convex quadratic in \(s\), 
    its minimum is attained at one or at two consecutive values of \(s\), leading to
    \(\multiplicity\) being either \(1\) or \(2\).
\end{proof}
\end{prop}

In particular, the case of Proposition \ref{prop:rlct_aw} is precisely the one studied by \cite{Aoyagi+Watanabe:2005}.  

\subsection{The Forward--Backward Covariance and the Geometry of Statistical Observability}

Positive definiteness as assumed in Proposition \ref{prop:rlct_aw} is a much stronger assumption than necessary.  Since $S = BA - B_0 A_0$ has rank at most $r + \rho$, the KL divergence only probes $\populationFB$ along a restricted family of low-rank directions.  This motivates a weaker nondegeneracy condition adapted to the geometry of LoRA.

From the eigendecomposition \(\populationFB = U\Lambda U^\transpose\) with a diagonal matrix 
\(\Lambda\in\RR^{D\times D}\) and an orthogonal matrix \(U\in\RR^{D\times D}\), we get 
\begin{equation}
\label{eq:kl_sandwich}
    K(A,B) = \frac12\norm{\Lambda^{\frac12}U^\transpose\vector{S}}^2.
\end{equation}
Let \(V_+\coloneqq\image\populationFB\) be the \emph{active subspace}, so that \(\RR^D = V_+\oplus\ker\populationFB\) is an orthogonal decomposition.  The orthogonal projection onto \(V_+\) is \(\Pi_+ \coloneqq U_+U_+^\transpose\), where \(U_+\) consists of the columns of \(U\) belonging to the positive eigenvalues.

Thus, the KL divergence depends only on the component of \(\vector{S}\) lying in \(V_+\).  Directions in \(\ker\populationFB\) are invisible to the locally linearized statistical model and perturbing the LoRA update along such a direction does not change the induced distribution.  The question we study is thus not whether \(\populationFB\) is nondegenerate on the entire ambient space, but rather, whether its kernel intersects the low-rank directions that can arise from LoRA.

As noted previously, \(\populationFB\) is a second-moment matrix, \(V_+\) is the span of the support of the distribution of \(\vector{\Phi(X)}\), and each
\(\Phi(X)=\backward(X)\forward(X)^\transpose\) has rank at most 1. Hence \(V_+\) is spanned by rank-one matrices, so not every linear subspace arises; for example, \(V_+=\langle I_2\rangle\subseteq\RR^{2\times2}\) does not.

Moreover, \(\rank\populationFB\leq\rank\forwardMoment\cdot\rank\backwardMoment\), where \(\forwardMoment=\mathbb E[\forward(X)\forward(X)^\transpose]\) and
\(\backwardMoment=\mathbb E[\backward(X)\backward(X)^\transpose]\). Indeed, if \(P_h\) and \(P_g\) are the orthogonal projections onto the images of
\(\forwardMoment\) and \(\backwardMoment\), then \(\forward(X)=P_h\forward(X)\) and \(\backward(X)=P_g\backward(X)\) almost surely, so
\(\vector{Z}\in\ker\populationFB\) whenever \(Z\in\RR^{m\times d}\) satisfies \(P_gZP_h=0\), and these \(Z\) form a subspace of codimension \(\rank P_g\cdot\rank P_h\).

The active-subspace decomposition identifies exactly which components of a perturbation contribute to the KL divergence.
To connect this geometry with the RLCT, we now compare the quadratic form induced by \(\populationFB\) 
to the Euclidean norm on its active subspace. 
Since the restriction of \(\populationFB\) to \(V_+\) is positive definite, its positive eigenvalues are bounded above 
and below away from zero. 
As a consequence, the KL divergence is equivalent to the squared norm of the projection of the update difference 
onto \(V_+\), up to positive multiplicative constants.  
We can adjust the sandwich bound from Proposition \ref{prop:rlct_aw} as follows.

\begin{lemma}\label{lem:sandwich}
    For the KL divergence \(K(A,B)\) as in \eqref{eq:kl_sandwich}, there exist positive real constants \(c_1,c_2\) 
    such that \[
        c_1\norm{\Pi_+\vector{S}}^2 \leq K(A,B) \leq c_2\norm{\Pi_+\vector{S}}^2.
    \]
\end{lemma}

For a possibly singular FB covariance, this is the analogue of the norm equivalence underlying the reduced-rank regression setting of \citet{Aoyagi+Watanabe:2005}.  The difference is that the Frobenius norm of the full matrix difference \(S\) is replaced by the norm of its projection onto the statistically observable subspace.

    \begin{proof}
        Write \(\Lambda = \diag(\sigma_1,\dots,\sigma_k,0,\dots,0)\) with \(\sigma_1\geq\dots\geq\sigma_k>0\), 
        so that \(U_+\in\RR^{D\times k}\) consists of the first \(k\) columns of \(U\). The last \(D-k\) entries of \(\Lambda^{\frac12}U^\transpose\vector{S}\)
        vanish, so with \(y\coloneqq U_+^\transpose\vector{S}\in\RR^k\) we get \[
            K(A,B) = \frac12\sum_{i=1}^k\sigma_iy_i^2,
            \quad\text{and hence}\quad
            \frac12\sigma_k\norm{y}^2\leq K(A,B)\leq\frac12\sigma_1\norm{y}^2.
        \] Since \(U_+\) has orthonormal columns, \(\norm{y}=\norm{U_+y}=\norm{\Pi_+\vector{S}}\). This proves the claim with 
        \(c_1=\frac12\sigma_k\) and \(c_2=\frac12\sigma_1\).
    \end{proof}

The map \((A,B)\mapsto BA\) is not injective: the factorization \(BA\) is non-identifiable and different parameter pairs \((A,B)\) can represent the same low-rank update.  This symmetry is naturally described by the following group action.

\begin{definition}[Gauge action]
\label{def:gauge_action}
The group \(\GenLinGrp{r}\) acts on the LoRA parameters by
\[
g\cdot(A,B)\coloneqq(gA,Bg^{-1}),
\qquad g\in\GenLinGrp{r}.
\]
We call this action the \emph{gauge action}, and its orbits \emph{gauge orbits}.
The fiber \(W_0\coloneqq\{\,(A,B)\mid BA=B_0A_0\,\}\) of \((A,B)\mapsto BA\) over the true update is the \emph{gauge fiber}.
\end{definition}

The product \(BA\) is invariant under the gauge action, since
\[
(Bg^{-1})(gA)=BA.
\]
Thus, \(K(A,B)\), which depends on the parameters only through \(BA\), turns out to be constant along gauge orbits.  This symmetry captures the intrinsic non-identifiability of the low-rank factorization.

The adjusted sandwich bound shows that, in the positive semidefinite case, the KL divergence is controlled by the projection of \(\vector{S}\) onto the active subspace rather than by the full product difference \(S=BA-B_0A_0\). This motivates the following map.

\begin{definition}[Active coordinate map]
The \emph{active coordinate map} is
\[
\activeCoord(A,B)
\coloneqq
\Pi_+\vector{BA-B_0A_0}.
\]
Its zero set is denoted by
\[
W_\Pi
\coloneqq
\left\{(A,B):
\Pi_+\vector{BA-B_0A_0}=0\right\}.
\]
\end{definition}

By Lemma \ref{lem:sandwich}, the zero set of \(K\) is
\(
W_\Pi=\{(A,B) \mid K(A,B)=0\}.
\)
Thus, \(W_\Pi\) is the set of LoRA parameters that are statistically indistinguishable from the ground truth under the locally linearized model.  At the level of matrix products, \(BA\) ranges over the determinantal variety of matrices of rank at most \(r\).

\begin{remark}\label{rem:determinantal}
    For \(k\geq0\), the matrices in \(\RR^{m\times d}\) of rank at most \(k\) form the \emph{determinantal variety} \(\detVar{\leq k}\), which is
    the common zero set of the \((k+1)\times(k+1)\) minors.
    For \(k\leq\min\{m,d\}\), the matrices of rank exactly \(k\) form a smooth submanifold \(\detVar{=k}\) of dimension \(k(m+d-k)\).
    Thus, \(\detVar{\leq k}\) is the disjoint union of the smooth strata \(\detVar{=j}\) for \(j\leq k\).
    For \(k<\min\{m,d\}\), the stratum \(\detVar{=k}\) is the smooth locus of \(\detVar{\leq k}\),
    and the lower strata \(\detVar{\leq k-1}\) form its singular locus.
    If \(Z=BA\) with \(B\in\RR^{m\times k}\) and \(A\in\RR^{k\times d}\) of rank \(k\), the tangent space of \(\detVar{=k}\) at \(Z\) is
    \(\{\delta_BA+B\delta_A\mid\delta_B\in\RR^{m\times k},\delta_A\in\RR^{k\times d}\}\).
\end{remark}

Like \(K\), the map \(\activeCoord\) depends on \((A,B)\) only through \(BA\), and \(W_\Pi\) is the union of the fibers of
\((A,B)\mapsto BA\) over the matrices \(Y\) with \(\vector{Y-B_0A_0}\in\ker\Pi_+\).
Since \(BA\) ranges over \(\detVar{\leq r}\), only the \(Y\) in the linear section \(\linearSection\coloneqq(B_0A_0+\ker\Pi_+)\cap\detVar{\leq r}\)
have a nonempty fiber, so \(W_\Pi\) is the preimage of \(\linearSection\) under \((A,B)\mapsto BA\). Each fiber is a union of gauge orbits.

If \(\rho=r\), the gauge fiber \(W_0\) is the gauge orbit of \((A_0,B_0)\) and \(\GenLinGrp{r}\) acts on it freely: for \((A,B)\in W_0\), the rows of
\(A\) and of \(A_0\) span the row space of \(B_0A_0\), so \(A=gA_0\) for a unique \(g\in\GenLinGrp{r}\), and \(BgA_0=B_0A_0\)
gives \(B=B_0g^{-1}\), since \(A_0\) has a right inverse.
If \(\rho<\min\{r,d\}\), it is not a single orbit: writing \(B_0A_0=UV\) with \(U\in\RR^{m\times\rho}\) and \(V\in\RR^{\rho\times d}\), it contains
\(\bigl(\begin{psmallmatrix}V\\Z\end{psmallmatrix},\begin{pmatrix}U&0\end{pmatrix}\bigr)\) for every \(Z\in\RR^{(r-\rho)\times d}\),
and the rank of \(\begin{psmallmatrix}V\\Z\end{psmallmatrix}\), which the gauge action preserves, is not constant in \(Z\).

By Lemma \ref{lem:sandwich} and \cite[Cor.~3.5]{MR3614481}, \(K\) and the components of \(\activeCoord\) have the same RLCT on any parameter domain.
It is the minimum of the local RLCTs of \(\activeCoord\) at the points of \(W_\Pi\) in that domain.

\subsection{The Nondegenerate Regime}
\label{sec:nondegenerate}

The geometry developed above allows the FB covariance to be singular in the ambient space: directions in \(\ker\Pi_+\) are statistically unobservable.  For LoRA, however, ambient degeneracy, by itself, is not the relevant notion of degeneracy.  The model can only explore differences \(S = BA - B_0A_0\) of rank at most \(r+\rho\).  So, even when \(\ker\Pi_+\) is nontrivial, it may be the case that there is no direction that can arise as such a low-rank difference.  In this case, projection onto the active subspace loses no information that would be important in distinguishing LoRA updates and the statistical geometry remains equivalent to that of reduced-rank regression.  We refer to this as the \emph{nondegenerate regime}.

The bounds of Lemma \ref{lem:sandwich} are in terms of \(\Pi_+\vector{S}\), which ignores the component of \(S\) in \(\ker\Pi_+\).
Yet, in some cases, the projection does not matter: \(K\) is still comparable to the full error \(\frobnorm{S}^2\), as in the setting of \citet{Aoyagi+Watanabe:2005}.
One such case is if \(\ker\Pi_+\) contains no nonzero matrix of rank at most \(r+\rho\), since \(S\) has no larger rank.

\begin{prop}\label{prop:aw-transfer}
    Let \(\rho=\rank B_0A_0\) and suppose that \(\ker\Pi_+\) contains no nonzero matrix of rank at most \(r+\rho\).
    Then there are \(c,c'>0\) with
    \[
        c\frobnorm{BA-B_0A_0}^2\leq K(A,B)\leq c'\frobnorm{BA-B_0A_0}^2
    \]
    for all \((A,B)\). In particular, Proposition \ref{prop:rlct_aw} applies to \(K\).
    \begin{proof}
        The set of matrices of rank at most \(r+\rho\) is a closed cone. Its intersection with the unit sphere is compact and misses \(\ker\Pi_+\),
        so there is \(c_0>0\) with \(\norm{\Pi_+\vector{Z}}\geq c_0\frobnorm{Z}\) for every \(Z\in\RR^{m\times d}\) of rank at most \(r+\rho\).
        Since \(\rank(BA-B_0A_0)\leq r+\rho\), Lemma \ref{lem:sandwich} gives the bounds with \(c=c_1c_0^2\) and \(c'=c_2\).
        The last claim then follows from Proposition \ref{prop:rlct_aw}.
    \end{proof}
\end{prop}

Another case in which we recover the regime from \citet{Aoyagi+Watanabe:2005} is when
the forward and backward signals are independent.
Then the second moment of the FB feature splits into the second moments of its two factors.

\begin{cor}\label{cor:kronecker-case}
    Suppose the input activation $\forward(X)$ and the backpropagated sensitivity $\backward(X)$ are independent
    and their second moments $\backwardMoment=\mathbb{E}[\backward(X)\backward(X)^{\transpose}]$ 
    and $\forwardMoment=\mathbb{E}[\forward(X)\forward(X)^{\transpose}]$ are positive definite.
    Then $\populationFB$ is positive definite and Proposition \ref{prop:rlct_aw} applies.
\end{cor}

    \begin{proof}
        Since \(\vector{\backward(X)\forward(X)^\transpose}=\forward(X)\otimes\backward(X)\), independence gives
        \[
            \populationFB=\mathbb{E}\left[\forward(X)\forward(X)^\transpose\otimes\backward(X)\backward(X)^\transpose\right]
            =\forwardMoment\otimes\backwardMoment.
        \]
        The eigenvalues of a Kronecker product are the products of the eigenvalues of its factors,
        so \(\populationFB\) is positive definite.
        This is the hypothesis of Proposition \ref{prop:rlct_aw}.
    \end{proof}

If \(\ker\Pi_+\) contains no nonzero matrix of rank at most \(2r\),
then Proposition \ref{prop:aw-transfer} applies to every \(B_0A_0\) of rank \(\rho\leq r\), since \(r+\rho\leq 2r\).
This condition says that the model distinguishes updates of rank at most \(r\): no two distinct matrices
\(Z_1,Z_2\in\RR^{m\times d}\) of rank at most \(r\) satisfy \[
    \frobprod{\Phi(X),Z_1}=\frobprod{\Phi(X),Z_2}
\] almost surely. Indeed, \(\vector{W}^\transpose\populationFB\vector{W}=\mathbb{E}\frobprod{\Phi(X),W}^2\), so \(W\in\RR^{m\times d}\) lies in \(\ker\Pi_+\) if and only if
\(\frobprod{\Phi(X),W}=0\) almost surely. Moreover, \(Z_1-Z_2\) has rank at most \(2r\), and every matrix of rank at most \(2r\) is such a difference.

The same argument gives the uniqueness of low-rank solutions of linear systems under a restricted isometry condition \cite[Theorem~3.2]{Recht.Fazel.Parrilo:2010}.
Since the set of matrices of rank at most \(2r\) and Frobenius norm \(1\) is compact, the condition is equivalent to \[
    \inf\{\vector{Z}^\transpose\populationFB\vector{Z}\mid Z\in\RR^{m\times d},\rank Z\leq 2r,\frobnorm{Z}=1\}>0.
\]

If \(\populationFB\) is singular, the hypothesis of Proposition \ref{prop:aw-transfer} requires \(r+\rho<\min\{m,d\}\),
since otherwise every matrix has rank at most \(r+\rho\).

Overall, the results here identify the regime in which degeneracy of the FB covariance does not influence the learning geometry of LoRA.  The covariance need not be positive definite on the ambient space: it is sufficient that its kernel contains no nonzero low-rank direction that can arise as a difference between admissible LoRA updates.  In this case, projection onto the active subspace preserves the relevant low-rank geometry and the KL divergence remains equivalent to the reduced-rank regression loss and the classical RLCT of \citet{Aoyagi+Watanabe:2005} therefore captures the learning complexity.

Outside of this regime, however, the kernel of the FB covariance intersects directions accessible to the low-rank model.  The zero set \(W_\Pi\) may then be strictly larger than the gauge fiber \(W_0\) and the reduced-rank regression RLCT no longer follows from the preceding comparison.  Understanding the learning complexity in this case requires a more in-depth analysis of the geometry of the active coordinate map, which is the problem that will be studied in the following section.

\section{Local Singular Geometry}
\label{sec:local}
We now study the local geometry of the model when the nondegeneracy condition of \Cref{sec:nondegenerate} fails.  Recall that by Lemma \ref{lem:sandwich}, the KL divergence is comparable to \(\norm{\activeCoord}^2\), where
\[
\activeCoord(A,B)
=
\Pi_+\vector{BA-B_0A_0},
\]
and its zero set \(W_\Pi\) consists of the parameters that are statistically indistinguishable from the ground truth under the locally linearized model.  When \(\ker\Pi_+\) intersects the relevant low-rank directions, this zero set can contain additional geometry beyond the intrinsic non-identifiability generated by the LoRA factorization.

Our aim in this section is to characterize this additional geometry locally around a point of \(W_\Pi\).  The main idea is to separate directions that are already statistically observable to the first order from the remaining directions in which the active coordinate map is singular.  The former component contributes a regular dimension-counting term to the RLCT, while the latter component determines a residual singularity.  The contact-equivalence and splitting results of \Cref{sec:contact_splitting} provide the tools needed for us to make this decomposition rigorous.

\subsection{Active Subspaces and the Local Product Geometry}

By Lemma \ref{lem:sandwich}, the local threshold of the model is controlled by the active coordinate map \(\activeCoord\), while Proposition \ref{prop:aw-transfer} determines the case where \(\ker\Pi_+\) contains no nonzero matrix of rank at most \(r+\rho\).  We now consider the general case.  Here, the geometry depends not only on the rank of the LoRA update, but also on the position of the active subspace \(V_+=\im \populationFB\) relative to the low-rank matrix geometry.

We therefore begin by examining the possible active subspaces of the FB covariance and their interaction with low-rank matrix directions.  This provides the geometric setting for our local analysis.  In particular, even the linear behavior of \(\activeCoord\) is not determined by ranks alone, as will be quantified by the dimension \(\kappa\) below.

We therefore begin by computing the local threshold at equal-rank points \(X'=(A',B')\in W_\Pi\), that is, points satisfying \(\rank A'=\rank B'=\rank B'A'\); further on in \Cref{sec:global}, we will show that these points suffice to determine the global threshold.  For this, we start with the unprojected parameter map, \[
    \varphi
    \colon
    \RR^{r\times d} \times \RR^{m\times r} \to \RR^D,\quad
    (A,B)\mapsto\vector{BA-B_0A_0}.
\] In a neighborhood of \(X'\), that is for \((A,B) = (A'+\delta_A,B'+\delta_B)\), we have \[
    \varphi(A,B)=\vector{B'A'-B_0A_0}+\vector{B'\delta_A + \delta_BA'}+\vector{\delta_B\delta_A}.
\] Write \(\delta=(\delta_A,\delta_B)\). Then the linear term is the linearization \[
    M\delta \coloneqq \differential\varphi(\delta_A,\delta_B) = \vector{B'\delta_A + \delta_BA'},
\] which we write as \(M_{X'}\) when the base point matters. 
By \(\vector{XYZ} = (Z^\transpose\otimes X)\vector{Y}\), we can represent this by the matrix
\[
    M\delta
    = (I_d\otimes B')\vector{\delta_A}
    + (A'^\transpose\otimes I_m)\vector{\delta_B}.
\] Since \(X'\in W_\Pi\), the constant term of \(\varphi\) lies in \(\ker\Pi_+\), and therefore \[
    \activeCoord(A,B) = \Pi_+M\delta + \Pi_+\vector{\delta_B\delta_A}.
\]

Before accounting for the projection onto the active subspace, we first determine the local geometry of the unprojected product map.  At an equal-rank point, this geometry depends only on the common rank \(\rho'\).  The following proposition computes the dimension of the kernel and image of its linearization \(M_{X'}\) and so it gives us the number of independent first-order directions in which the product \(BA\) can vary.  The effect of the FB covariance can then be understood as removing from these directions the ones that lie in \(\ker\Pi_+\).

\begin{prop}\label{prop:rank-M}
    Let \(X=(A,B)\in W_\Pi\) be an equal-rank point with rank \(\rho'\), then
    \[
        \dim\ker{M}
        =
        \rho'^2 + (r-\rho')(m+d)
        \text{ and }
        \rank{M}
        =
        \rho'(m+d-\rho').
    \]
\end{prop}

    \begin{proof}
        Due to \(\rank{A} = \rank{BA} = \rho'\) and \(\dim\ker{B} = r - \rho'\), we have 
        \(\image{A}\cap\ker{B} = 0\) and we can find a decomposition 
        \(\RR^r = \image{A}\oplus\ker{B}\). The gauge action satisfies \(M_{g\cdot X}(g\cdot\delta)=M_X(\delta)\), so a change of basis in \(\RR^r\) does not affect
        the rank or the kernel dimension of \(M\). Thus after a change of basis,
        write
        \(A = \begin{psmallmatrix} A_1 \\ 0 \end{psmallmatrix}\) and
        \(B = \begin{psmallmatrix} B_1 & 0 \end{psmallmatrix}\)
        where \(A_1\) and \(B_1\) have both rank \(\rho'\).

        Applying the same block decomposition to \(
            \delta_A 
            = 
            \begin{psmallmatrix} \delta_{A,1} \\ \delta_{A,2} \end{psmallmatrix}
        \) and \(
            \delta_B
            = 
            \begin{psmallmatrix} \delta_{B,1} & \delta_{B,2} \end{psmallmatrix}
        \) we get \[
            M(\delta_A,\delta_B) 
            = B_1\delta_{A,1} + \delta_{B,1}A_1
            = M_1(\delta_{A,1},\delta_{B,1})
        \] where \(M_1\) is the restriction of \(M\) to the corresponding coordinates. 
        
        In particular, if \(P\) is the projection onto those
        coordinates, then \(M = M_1\circ P\) and \[
            \ker{M}
            = P^{-1}(\ker{M_1})
            = \ker{M_1} \oplus \ker{P}
        \] due to the block decomposition of \(\delta_A\) and \(\delta_B\).

        Since \(A_1\) has rank \(\rho'\), it has a right inverse \(A_1^{\mathrm{r}}\), and since \(B_1\) has rank \(\rho'\), it has a left inverse \(B_1^{\mathrm{l}}\).
        Suppose that \(M_1(\delta_{A,1},\delta_{B,1}) = B_1\delta_{A,1} + \delta_{B,1}A_1 = 0\). Multiplying from the right by \(A_1^{\mathrm{r}}\) gives
        \(\delta_{B,1} = -B_1\xi\) for \(\xi\coloneqq\delta_{A,1}A_1^{\mathrm{r}}\in\RR^{\rho'\times\rho'}\). Then 
        \(B_1(\delta_{A,1}-\xi A_1) = 0\), and multiplying from the left by \(B_1^{\mathrm{l}}\) gives \(\delta_{A,1} = \xi A_1\).
        Conversely, every pair \((\xi A_1,-B_1\xi)\) lies in \(\ker M_1\), and \(\xi\mapsto(\xi A_1,-B_1\xi)\) is injective since
        \(A_1\) has a right inverse. Thus \(\ker M_1\cong\RR^{\rho'\times\rho'}\) and \(\dim\ker{M_1} = \rho'^2\).
        
        On the other hand, by inspection we get \[
            \ker{P} = \RR^{(r-\rho')\times d}\oplus\RR^{m\times(r-\rho')}.
        \] This implies the dimension count for \(\ker{M}\), and the rank count follows
        by rank--nullity, as \(M\) is defined on a space of dimension \(r(m+d)\).
    \end{proof}

The preceding proposition describes the first-order geometry of the LoRA product map independently of the FB covariance.  The active coordinate map, however, observes these variations only after projection onto \(V_+\).  Thus, the relevant linearization for the statistical model is not \(M_{X'}\) itself, but rather
$$
\differential\activeCoord_{X'}=\Pi_+M_{X'}.
$$
As a consequence, directions in \(\image M_{X'}\cap\ker\Pi_+\) correspond to first-order variations of the LoRA update that are invisible to the locally linearized statistical model.  We quantify this loss of observable tangent directions by
\[
\kappa(X')
\coloneqq
\dim\bigl(\image M_{X'}\cap\ker\Pi_+\bigr).
\]
At an equal-rank point, Proposition \ref{prop:rank-M} then allows the rank of \(\differential\activeCoord_{X'}\) to be expressed directly in terms of the product rank and \(\kappa(X')\).

\subsection{Local RLCT and the Residual Singularity}

We now translate the first-order geometry of the active coordinate map into a local RLCT.  The observable tangent directions just identified contribute a regular component to the local threshold, while any remaining singular behavior is captured by a residual function.  The splitting result of Lemma \ref{lem:splitting} makes this decomposition precise.

By Lemma \ref{lem:splitting}, \(\activeCoord\) splits at \(X'\) into \(s\) regular coordinates and a residual function \(\chi\).
At an equal-rank point, \(\kappa(X')\) determines \(s\), so the local threshold is governed by \(\kappa(X')\) and \(\chi\).
To compute the local threshold at \(X'\), we combine Corollary \ref{cor:splitting-rlct} with Proposition \ref{prop:rank-M} as follows.

\begin{theorem}\label{thm:reduction}
    Let \(X'=(A',B')\in W_\Pi\), let \(s=\rank\differential \activeCoord_{X'}\) and let \(\chi\) be 
    a real-analytic function as in Lemma \ref{lem:splitting} for \(\activeCoord\) at \(X'\).
    \begin{enumerate}
        \item The local threshold \(\rlct_{X'}(\activeCoord)=(\lambda_{X'},\multiplicity_{X'})\) is given by 
            \[
                \lambda_{X'} = \frac{s}{2} + \lambda_\chi
                \quad\text{and}\quad
                \multiplicity_{X'} = \multiplicity_\chi,
            \] where \((\lambda_\chi,\multiplicity_\chi)=\rlct_0(\chi_1,\dots,\chi_{D-s})\), 
            or \((\lambda_\chi,\multiplicity_\chi)=(0,1)\) if \(\chi\equiv0\).
        \item If moreover \(X'\) is an equal-rank point with rank \(\rho'\),
            then \(s=\rho'(m+d-\rho') - \kappa(X')\). In particular, \[
                \lambda_{X'} = \frac{ \rho'(m+d-\rho') - \kappa(X')}{2} + \lambda_\chi.
            \]
    \end{enumerate}
\end{theorem}

    \begin{proof}
        The first statement is Corollary \ref{cor:splitting-rlct} for \(F=\activeCoord\), as \(\activeCoord(X')=0\).
        For the second statement, since \(\differential\activeCoord_{X'} = \Pi_+M_{X'}\), we have
        \(\ker(\Pi_+\vert_{\image M_{X'}}) = \image{M_{X'}}\cap\ker\Pi_+\), 
        and by rank--nullity we obtain \[
             s = \rank\Pi_+ M_{X'} = \rank M_{X'} - \kappa(X') = \rho'(m+d-\rho') - \kappa(X') ,
        \] where the last step is Proposition \ref{prop:rank-M}. This implies the expression for \(\lambda_{X'}\).
    \end{proof}

At this point, \Cref{thm:reduction} leaves \(\lambda_\chi\) open. We first ask when \(\chi\equiv0\), in which case \(\lambda_{X'}=\frac s2\).
The simplest case is when \(\rho'=r\) and \(\kappa(X')=0\), that is, \(\Pi_+\) loses no first-order change of \(BA\) at \(X'\).
This can be checked on \(\populationFB\): \(\kappa(X')=0\) if and only if \(\populationFB\) is positive
definite on \(\image M_{X'}\), since \(\ker\Pi_+=\ker\populationFB\).
By rank--nullity, \(\kappa(X')\geq\rank M_{X'}-\rank\populationFB\), so \(\kappa(X')=0\) is impossible if \(\rank\populationFB<\rank M_{X'}\).
In this case, the following result recovers an instance of Proposition \ref{prop:rlct_aw} for \(\rho'=r\).

\begin{cor}\label{cor:clean-case}
    Let \(X'=(A',B')\in W_\Pi\) be an equal-rank point with rank \(r\) and \(\kappa(X')=0\). Then \[
        \lambda_{X'}=\frac{r(m+d-r)}{2}\quad\text{and}\quad\multiplicity_{X'}=1.
    \]
    \begin{proof}
        Here \(\rho'=r\) and \(\kappa(X')=0\), so \(s=r(m+d-r)\) by \Cref{thm:reduction}, and the \(y\)-space of Lemma \ref{lem:splitting}
        has dimension \(r(m+d)-s=r^2\). Since \(A'\) has rank \(r\), the stabilizer of \(X'\) in \(\GenLinGrp{r}\) 
        is trivial, so near \(X'\) the gauge orbit of \(X'\) is a submanifold of dimension \(r^2\).
        It consists of points with \(BA=B'A'\), so it lies in \(W_\Pi\), and thus in \(\{x=0\}\) in the coordinates 
        of Lemma \ref{lem:splitting}. Hence, near \(X'\), it is an open subset of the \(y\)-space containing \(0\), 
        and \(\chi\) vanishes on it, so \(\chi\equiv0\) near \(0\). 
        By Corollary \ref{cor:splitting-rlct}, we obtain \((\lambda_{X'},\multiplicity_{X'})=(\frac{s}{2},1)\).
    \end{proof}
\end{cor}

Beyond this case, \(\chi\) need not vanish.
In the proof of Corollary \ref{cor:clean-case}, \(\rho'=r\) and \(\kappa(X')=0\) are used only to see that \(W_\Pi\) has dimension \(r(m+d)-s\) at \(X'\), so at a general equal-rank point we look at this dimension.

\begin{remark}\label{rem:vanishing-residual}
    Let \(X'\in W_\Pi\) be an equal-rank point with rank \(\rho'\) and let \(s=\rho'(m+d-\rho')-\kappa(X')\).
    In the coordinates of Lemma \ref{lem:splitting}, \(W_\Pi\) is \(\{x=0,\chi(y)=0\}\) near \(X'\), so \(\chi\equiv0\) in \Cref{thm:reduction} if and only if the
    dimension of \(W_\Pi\) at \(X'\) equals \(r(m+d)-s\). In this case, \(\rlct_{X'}(\activeCoord)=(\tfrac s2,1)\), and \(K\) is a Morse--Bott function near \(X'\).
    For \(\rho'=r\), the product map is a submersion onto \(\detVar{=r}\) near \(X'\) whose fibres are gauge orbits of dimension \(r^2\),
    so the condition says that \(\linearSection\) has dimension \(\kappa(X')\) at \(B'A'\).
\end{remark}

For \(\rho'=r\), the linear section \(\linearSection\) has dimension at most \(\kappa(X')\) at \(B'A'\), since its tangent space there lies in \(\ker\Pi_+\cap\image M_{X'}\), and \(\image M_{X'}\) is the tangent space of \(\detVar{=r}\).
If \(\kappa(X')=0\), the condition of Remark \ref{rem:vanishing-residual} therefore holds, which is Corollary \ref{cor:clean-case}.
Otherwise \(\chi\not\equiv0\) exactly when this bound is not attained, and both cases occur in the following example.

\begin{example}\label{ex:two-by-two}
    Let \(m=d=2\), \(r=\rho=1\), \(A_0=(1\ 0)\) and \(B_0=(1\ 0)^\transpose\), and write \(A=(a_1\ a_2)\), \(B=(b_1\ b_2)^\transpose\) and
    \(u\coloneqq a_1b_1-1\). Then
    \[
        BA-B_0A_0=\begin{pmatrix}u&b_1a_2\\b_2a_1&b_2a_2\end{pmatrix}
    \]
    and \(\image M=\{Z\mid Z_{22}=0\}\). For a set \(J\) of positions, let \(\Pi_+\) be the orthogonal projection with kernel spanned by
    the \(E_{ij}\), \((i,j)\in J\). Near \((A_0,B_0)\) we get the values in \Cref{tab:two-by-two}.
    \begin{table}[ht]
    \centering
    \caption{The local data of Example \ref{ex:two-by-two} at \((A_0,B_0)\), for three choices of \(J\).}\label{tab:two-by-two}
    \begin{tabular}{ccccc}
        \toprule
        \(J\) & \(\kappa\) & \(\norm{\activeCoord}^2\asymp\) & \((\lambda,\multiplicity)\) & \Cref{thm:reduction} \\ \midrule
        \(\emptyset\) & \(0\) & \(u^2+a_2^2+b_2^2\) & \((\frac32,1)\) & \(s=3\), \(\chi\equiv0\) \\
        \(\{(1,2)\}\) & \(1\) & \(u^2+b_2^2(a_1^2+a_2^2)\) & \((1,1)\) & \(s=2\), \(\chi\equiv0\) \\
        \(\{(1,2),(2,1)\}\) & \(2\) & \(u^2+(a_2b_2)^2\) & \((1,2)\) & \(s=1\), \(\chi=a_2b_2\) \\\bottomrule
    \end{tabular}
    \end{table}
    The second and third projections are those of an FB covariance whose features are supported on \(E_{11},E_{21},E_{22}\), respectively
    on \(E_{11},E_{22}\). In the third case \(\rho=r\), and still \(\chi\not\equiv0\).
    One way to obtain such features is to split the input domain \(\mathcal{X}\) into regimes \(\mathcal{X}_i\) on which \(\backward(x)\) and \(\forward(x)\) have different nonzero coordinates.
    On a regime, \(\Phi(x)\) is supported on the entries \((i,j)\) for which the \(i\)th coordinate of \(\backward(x)\) and the \(j\)th coordinate of \(\forward(x)\) are both nonzero,
    and the union of these supports is the complement of \(J\).
    \Cref{fig:bipartitegraph_two-by-two} shows these supports and their union as bipartite graphs.
\end{example}

\definecolor{regA}{gray}{0.10}       
\definecolor{regB}{gray}{0.50}       
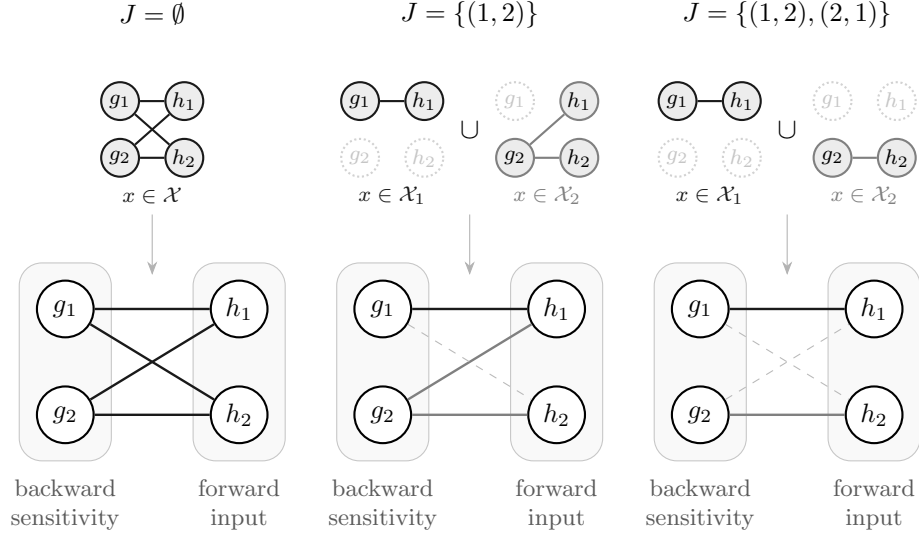
\begin{figure}[h]
\centering
\begin{tikzpicture}[
  unit/.style={circle, draw, thick, fill=white, minimum size=7.5mm, inner sep=0pt},
  mini/.style={circle, draw, thick, minimum size=5mm, inner sep=0pt, font=\scriptsize},
  off/.style={mini, draw=gray!35, densely dotted, text=gray!45, fill=white},
  group/.style={draw=gray!45, fill=gray!5, rounded corners=3mm, inner sep=2.2mm},
  unseen/.style={thin, dashed, gray!55},
  every node/.style={font=\small}]
 
\newcommand{\regime}[9]{
  \begin{scope}[shift={(#1,#2)}]
    \node[#3] (a1) at (0, 0.75) {$g_1$};  \node[#4] (a2) at (0, 0) {$g_2$};
    \node[#5] (b1) at (0.85, 0.75) {$h_1$}; \node[#6] (b2) at (0.85, 0) {$h_2$};
    \foreach \u/\v in {#8} \draw[#7, thick] (a\u) -- (b\v);
    \node[font=\scriptsize, text=#7] at (0.425, -0.5) {#9};
  \end{scope}}
\newcommand{\union}[1]{
  \begin{scope}[shift={(#1,-3.1)}]
    \node[unit] (g1) at (0, 1.4) {$g_1$}; \node[unit] (g2) at (0, 0) {$g_2$};
    \node[unit] (h1) at (2.3, 1.4) {$h_1$}; \node[unit] (h2) at (2.3, 0) {$h_2$};
    \begin{scope}[on background layer]
      \node[group, fit=(g1)(g2)] (G) {}; \node[group, fit=(h1)(h2)] (H) {};
    \end{scope}
    \node[gray!70!black, font=\footnotesize, align=center, below=1mm] at (G.south) {backward\\sensitivity};
    \node[gray!70!black, font=\footnotesize, align=center, below=1mm] at (H.south) {forward\\input};
  \end{scope}}
 
\node at (1.15, 2.2) {$J=\emptyset$};
\regime{0.725}{0.3}{mini, draw=regA, fill=regA!8}{mini, draw=regA, fill=regA!8}{mini, draw=regA, fill=regA!8}{mini, draw=regA, fill=regA!8}{regA}{1/1,1/2,2/1,2/2}{$x\in\mathcal X$}
\union{0}
\foreach \u/\v in {1/1,1/2,2/1,2/2} \draw[regA, semithick, line width=0.9pt] (g\u) -- (h\v);
 
\begin{scope}[xshift=4.2cm]
\node at (1.15, 2.2) {$J=\{(1,2)\}$};
\regime{-0.3}{0.3}{mini, draw=regA, fill=regA!8}{off}{mini, draw=regA, fill=regA!8}{off}{regA}{1/1}{$x\in\mathcal{X}_{1}$}
\node[font=\normalsize] at (1.15, 0.67) {$\cup$};
\regime{1.75}{0.3}{off}{mini, draw=regB, fill=regB!15}{mini, draw=regB, fill=regB!15}{mini, draw=regB, fill=regB!15}{regB}{2/1,2/2}{$x\in\mathcal{X}_{2}$}
\union{0}
\draw[regA, semithick, line width=0.9pt] (g1) -- (h1);
\draw[regB, semithick, line width=0.9pt] (g2) -- (h1); \draw[regB, semithick, line width=0.9pt] (g2) -- (h2);
\draw[unseen] (g1) -- (h2);
\end{scope}
 
\begin{scope}[xshift=8.4cm]
\node at (1.15, 2.2) {$J=\{(1,2),(2,1)\}$};
\regime{-0.3}{0.3}{mini, draw=regA, fill=regA!8}{off}{mini, draw=regA, fill=regA!8}{off}{regA}{1/1}{$x\in\mathcal{X}_{1}$}
\node[font=\normalsize] at (1.15, 0.67) {$\cup$};
\regime{1.75}{0.3}{off}{mini, draw=regB, fill=regB!15}{off}{mini, draw=regB, fill=regB!15}{regB}{2/2}{$x\in\mathcal{X}_{2}$}
\union{0}
\draw[regA, semithick, line width=0.9pt] (g1) -- (h1);
\draw[regB, semithick, line width=0.9pt] (g2) -- (h2);
\draw[unseen] (g1) -- (h2); \draw[unseen] (g2) -- (h1);
\end{scope}
\foreach \xs in {0, 4.2, 8.4} \draw[-{Stealth[length=1.8mm]}, gray!60] (\xs+1.15, -0.45) -- (\xs+1.15, -1.25);
\end{tikzpicture}
\caption{Bipartite graphs of the three cases of Example \ref{ex:two-by-two}, one for each row of \Cref{tab:two-by-two}.
Top: on each input regime \(\mathcal{X}_i\), the FB feature joins the nonzero coordinates of \(g(x)\) to those of \(h(x)\); dotted coordinates vanish on \(\mathcal{X}_i\).
Bottom: the union over the regimes, with the entries in \(J\) dashed.}
\label{fig:bipartitegraph_two-by-two}
\end{figure}

In Example \ref{ex:two-by-two}, the section \(\linearSection\) is the point \(E_{11}\) for \(J=\emptyset\) and 
the line \(E_{11}+\RR E_{12}\) for \(J=\{(1,2)\}\).
Both are smooth of respective dimension \(\kappa\), so the bound is attained and \(\chi\equiv0\) by Remark \ref{rem:vanishing-residual}.

For \(J=\{(1,2),(2,1)\}\), \(\linearSection\) is the union of the lines \(E_{11}+\RR E_{12}\) and \(E_{11}+\RR E_{21}\), 
which cross at \(E_{11}\).
Here, the section is one-dimensional, but its tangent space at \(E_{11}\) is \(\ker\Pi_+\), which has dimension \(\kappa\), 
so the bound is not attained and \(\chi\not\equiv0\).

In all three cases, \(\ker\Pi_+\) is spanned by matrix units, so \(\Pi_+\) keeps exactly the entries outside \(J\).
For such projections, \(s\) at generic points comes from low-rank matrix completion.

\begin{remark}\label{rem:completion}
    Let \(J\subseteq[m]\times[d]\), let \(\Pi_+\) be the orthogonal projection with kernel spanned by the \(E_{ij}\), 
    \((i,j)\in J\), as in \Cref{ex:two-by-two}, and let \(F\) be the complement of \(J\). 
    Let \(X'=(A',B')\in W_\Pi\) be an equal-rank point with rank \(r\), and let \(Y=B'A'\).
    
    Then \(\image M_{X'}=T_Y\detVar{=r}\) by Remark \ref{rem:determinantal}, so \(s=\rank \differential\activeCoord_{X'}\) 
    is the rank at \(Y\) of the differential of the coordinate projection \(\pi_F\colon\detVar{=r}\to\RR^F\). 
    Its maximum over \(\detVar{=r}\) is \(\rank_r(F)\) of \(F\) in the rank-\(r\) determinantal matroid
    \cite[Lem.~8]{KiralyTheranTomioka2015}, which is defined over \(\CC\) and has the same rank on \(\detVar{=r}\), 
    since the real matrices of rank \(r\) are Zariski dense.
    
    Let \(\Sigma_F\) be the proper Zariski-closed subset of \(\detVar{=r}\) where the rank of \(\differential\pi_F\) 
    is smaller.
    If \(Y\notin\Sigma_F\), then \(\rank \differential\activeCoord\) is constant near \(X'\), 
    so \(W_\Pi\) is a submanifold of dimension \(r(m+d)-s\) near \(X'\). 
    Since \(s=\rank_r(F)=r(m+d-r)-\kappa(X')\), Remark \ref{rem:vanishing-residual} gives
    \[
        \rlct_{X'}(\activeCoord)=\Bigl(\frac{\rank_r(F)}{2},1\Bigr),\qquad \kappa(X')=r(m+d-r)-\rank_r(F).
    \]
    This applies at \((A_0,B_0)\) if \(B_0A_0\) has rank \(r\) and lies outside \(\Sigma_F\), in particular for generic \(B_0A_0\) of rank \(r\).
\end{remark}

Thus, for these projections and \(\rho'=r\), a nonzero residual can only occur at products in \(\Sigma_F\).  In the rank-one case, the generic value of \(\kappa\) admits a simple graph-theoretic interpretation.

For \(r=1\), \(\rank_1(F)\) can be computed from a bipartite graph in the following way: For \(r=1\) and \(Y=pq^\transpose\) with no zero row and no zero column, \(\differential\pi_F\) at \(Y\) is, up to invertible diagonal scalings, the incidence map \((x,y)\mapsto(x_j+y_i)_{(i,j)\in F}\) of
    the bipartite graph on \([m]\sqcup[d]\) with edge set \(F\). So \(s=m+d-c\), where \(c\) is the number of connected components of this graph, and \(\kappa(X')=c-1\).
    This is \(\rank_1(F)\), since the rank-one determinantal matroid is the cycle matroid of this graph \cite[Thm.~34]{KiralyTheranTomioka2015}.
    
    For Example \ref{ex:two-by-two}, this graph consists of the solid edges in the bottom row of \Cref{fig:bipartitegraph_two-by-two}.
    In Example \ref{ex:two-by-two}, the graph is connected for \(J=\emptyset\) and \(J=\{(1,2)\}\), so \(\kappa(X')=0\) outside \(\Sigma_F\) in these two cases, and it has two components for \(J=\{(1,2),(2,1)\}\), so \(\kappa(X')=1\) outside \(\Sigma_F\) there.
    At \(Y=E_{11}\), the value of \(\kappa\) in \Cref{tab:two-by-two} is larger than \(c-1\) for \(J=\{(1,2)\}\) and \(J=\{(1,2),(2,1)\}\), so \(E_{11}\in\Sigma_F\) in these two cases.

\paragraph{The Residual Determinantal Model.}  By \Cref{thm:reduction}, the local threshold at an equal-rank point is known up to the threshold of the residual function \(\chi\), and by \Cref{sec:global}
these points determine the global threshold. What remains is therefore \(\chi\) at the equal-rank points over \(\linearSection\).
On a slice transverse to the gauge orbit of an equal-rank point with rank \(\rho'\), the residual variables form a determinantal model of rank at most \(r-\rho'\) 
in a linear subspace of \(\RR^{m\times d}\) of dimension \((m-\rho')(d-\rho')\) via \[(\delta_A,\delta_B)\mapsto\delta_B\delta_A.\]
To see this, we first bring the equal-rank point into a normal form and then pass to the slice.

\paragraph{Normalizing an Equal-Rank Point.}
Since \(B'A'-B_0A_0\in\ker\Pi_+\) for an equal-rank point \(X'=(A',B')\in W_\Pi\) with rank \(\rho'\), 
we have \(\activeCoord(A,B)=\Pi_+\vector{BA-B'A'}\), so we assume without loss of
generality that \((A_0,B_0)\) is an equal-rank point.

Then, we have  \(\RR^r=\image A_0\oplus\ker B_0\) 
as in the proof of Proposition \ref{prop:rank-M}.
So, after a change of basis \((A,B)\mapsto(gAG_2^\transpose,G_1Bg^{-1})\) by the gauge action 
together with invertible matrices \(G_1\) and \(G_2\),
we may assume \[
    A_0 = \begin{pmatrix}
        I_{\rho'} & 0 \\
        0 & 0
    \end{pmatrix}
    \quad\text{and}\quad
    B_0 = \begin{pmatrix}
        I_{\rho'} & 0 \\
        0 & 0
    \end{pmatrix}.
\] The change of basis sends \(BA\) to \(\tau(BA)\) for \(\tau(Z)\coloneqq G_1ZG_2^\transpose\).
Let \(\Pi_+'\) be the orthogonal projection with kernel \(\tau(\ker\Pi_+)\). The components of \(\Pi_+\vector{Z}\) and of \(\Pi_+'\vector{\tau(Z)}\) are linear in \(Z\) with the same kernel, so they generate the same ideal.
By Proposition \ref{prop:ideal-independence} and \cite[Prop.~3.8 and Lemma~2.4]{MR3614481}, the change of basis does not change the RLCT, and we may replace \(\Pi_+\) by \(\Pi_+'\).

Writing the first-order perturbations \(\delta_A\) and \(\delta_B\) in the corresponding block form then
allows us to express the linearization \(M\) as 
\begin{equation}\label{eq:normal-form-residual}
    M(\delta_A,\delta_B) 
    = B_0\delta_A + \delta_BA_0
    = \begin{pmatrix}
        a_{11} + b_{11} & a_{12} \\
        b_{21}          & 0
    \end{pmatrix}.
\end{equation} The coordinates \(a_{22}\) and \(b_{22}\) do not enter \(M\), but they are not gauge directions either, so for \(\rho'<r\) the kernel of \(M\) contains directions that are not tangent to the gauge orbit.

\paragraph{A Slice Transverse to the Gauge Orbit.}
We therefore use a slice whose directions are complementary to the tangent space of the orbit.
Let \(\gaugeSlice = (A_0,B_0)+\vec{\gaugeSlice}\) be the affine subspace through \((A_0,B_0)\) whose directions \((\delta_A,\delta_B)\) satisfy
\begin{equation}\label{eq:gauge-slice}
    a_{21}=0,\qquad b_{12}=0,\qquad a_{11}=b_{11}
\end{equation}
in the block coordinates given by the normal form.
For \(\xi\in\GenLinAlg{r}\) with blocks \(\xi_{ij}\), the tangent vector of the gauge orbit of \((A_0,B_0)\) in the direction \(\xi\) is \((\xi A_0,-B_0\xi)\), where
\begin{equation}\label{eq:gauge-tangent-vector}
    \xi A_0=\begin{pmatrix}\xi_{11}&0\\\xi_{21}&0\end{pmatrix},\qquad -B_0\xi=-\begin{pmatrix}\xi_{11}&\xi_{12}\\0&0\end{pmatrix}.
\end{equation} So the tangent space \(T\) of the orbit at \((A_0,B_0)\) consists of the directions with \(a_{11}=-b_{11}\) and \(a_{12}=a_{22}=b_{21}=b_{22}=0\), where \(a_{21}\) and \(b_{12}\) are free,
and \(\dim T=\rho'(2r-\rho')\).

\begin{prop}\label{lem:local-gauge-slice}
    Let \(\mathfrak{t}\coloneqq\{\xi\in\GenLinAlg{r}\mid\xi_{22}=0\}\), in block coordinates of sizes \(\rho'\) and \(r-\rho'\).
    \begin{enumerate}[label=(\arabic*)]
        \item \(T\cap\vec{\gaugeSlice}=0\).
        \item The map \(\Theta\colon\mathfrak{t}\times\vec{\gaugeSlice}\to\RR^{r\times d}\times\RR^{m\times r}\) 
            given by \((\xi,\eta)\mapsto\exp(\xi)\cdot\bigl((A_0,B_0)+\eta\bigr)\)
            is a local real-analytic diffeomorphism at \((0,0)\).
        \item The local RLCT of \(\activeCoord\) at \((A_0,B_0)\) equals that of \(\activeCoord\vert_{\gaugeSlice}\).
    \end{enumerate}
\end{prop}

    \begin{proof}
        By \eqref{eq:gauge-tangent-vector}, the map \(\xi\mapsto(\xi A_0,-B_0\xi)\) is injective on \(\mathfrak{t}\), since \((\xi A_0,-B_0\xi)\) determines \(\xi_{11},\xi_{12},\xi_{21}\),
        and \(\dim\mathfrak{t}=r^2-(r-\rho')^2=\dim T\).

        The three equations in \eqref{eq:gauge-slice} impose \(\rho'(r-\rho')\), \(\rho'(r-\rho')\) and \(\rho'^2\) independent conditions, in total \(\dim T\), so
        \(\dim T+\dim\vec{\gaugeSlice}=r(m+d)\). If \((\delta_A,\delta_B)\in T\cap\vec{\gaugeSlice}\), then \(a_{21}=b_{12}=0\), and \(a_{11}=b_{11}=-a_{11}\) forces
        \(a_{11}=b_{11}=0\); all other coordinates vanish on \(T\). Hence \(T\cap\vec{\gaugeSlice}=0\). The differential
        \(\differential\Theta_{(0,0)}(\xi,\eta)=(\xi A_0,-B_0\xi)+\eta\) is therefore injective, hence an isomorphism by the dimension count, and \(\Theta\) is a local
        real-analytic diffeomorphism by the inverse function theorem. Since \(BA\) is invariant under the gauge action, the function
        \(\activeCoord\circ\Theta(\xi,\eta)=\activeCoord\bigl((A_0,B_0)+\eta\bigr)\) does not depend on \(\xi\). Local RLCTs are invariant under
        real-analytic changes of variables, and integrating out \(\xi\) only changes the amplitude \cite[Prop.~3.8 and Lemma~2.4]{MR3614481}.
    \end{proof}

\begin{lemma}\label{lem:slice-linearization}
    Let \(\gaugeSlice_M\) and \(\gaugeSlice_{\mathrm{res}}\) be the subspaces of \(\vec{\gaugeSlice}\) on which \((a_{22},b_{22})\), respectively \((a_{11},a_{12},b_{21})\), vanish.
    \begin{enumerate}[label=(\arabic*)]
        \item \(\vec{\gaugeSlice}=\gaugeSlice_M\oplus\gaugeSlice_{\mathrm{res}}\). The map \(M\) restricts to an isomorphism \(M\vert_{\gaugeSlice_M}\colon\gaugeSlice_M\to\image{M}\),
            it vanishes on \(\gaugeSlice_{\mathrm{res}}\), and \(\gaugeSlice_{\mathrm{res}}\cong\ker M/T\).
        \item For \(h\in\GenLinGrp{r-\rho'}\), the element \(\diag(I_{\rho'},h)\) fixes \((A_0,B_0)\), preserves \(\gaugeSlice\), and acts on \(\gaugeSlice_{\mathrm{res}}\) by
            \((a_{22},b_{22})\mapsto(ha_{22},b_{22}h^{-1})\).
    \end{enumerate}
\end{lemma}

    \begin{proof}
        (1) The perturbations in \(\vec{\gaugeSlice}\) satisfy \(a_{21}=b_{12}=0\) and \(b_{11}=a_{11}\), so their free coordinates are
        \(a_{11},a_{12},b_{21}\) and \(a_{22},b_{22}\). Hence \(\vec{\gaugeSlice}=\gaugeSlice_M\oplus\gaugeSlice_{\mathrm{res}}\), where \(\gaugeSlice_M\)
        is the subspace with \(a_{22}=b_{22}=0\) and \(\gaugeSlice_{\mathrm{res}}\) the subspace with \(a_{11}=a_{12}=b_{21}=0\).

        By eq.\ \eqref{eq:normal-form-residual}, \(M(\delta_A,\delta_B)=\begin{psmallmatrix}2a_{11}&a_{12}\\b_{21}&0\end{psmallmatrix}\) on \(\gaugeSlice_M\),
        and \(M=0\) on \(\gaugeSlice_{\mathrm{res}}\). So \(M\vert_{\gaugeSlice_M}\) is injective, and
        \(\dim\gaugeSlice_M=\rho'^2+\rho'(d-\rho')+\rho'(m-\rho')=\rho'(m+d-\rho')=\rank M\) by Proposition \ref{prop:rank-M}.
        Hence \(M\vert_{\gaugeSlice_M}\colon\gaugeSlice_M\to\image M\) is an isomorphism. 
        Also, \(\image M\) lies in the matrices with vanishing
        lower right block, which form a space of the same dimension \(\rho'(m+d-\rho')\), so they are equal.

        For the last claim, note that \(T\subseteq\ker M\), since \(BA\) is constant on the orbit, and \(\gaugeSlice_{\mathrm{res}}\subseteq\ker M\).
        By Lemma \ref{lem:local-gauge-slice}, \(T\cap\gaugeSlice_{\mathrm{res}}\subseteq T\cap\vec{\gaugeSlice}=0\). By Proposition \ref{prop:rank-M},
        \(\dim T+\dim\gaugeSlice_{\mathrm{res}}=\rho'(2r-\rho')+(r-\rho')(m+d-2\rho')=\rho'^2+(r-\rho')(m+d)=\dim\ker M\),
        so \(\ker M=T\oplus\gaugeSlice_{\mathrm{res}}\) and \(\gaugeSlice_{\mathrm{res}}\cong\ker M/T\).

        (2) Let \(g\coloneqq\diag(I_{\rho'},h)\). Then \(gA_0=A_0\) and \(B_0g^{-1}=B_0\), so \(g\) maps \((A_0+\delta_A,B_0+\delta_B)\) to \((A_0+g\delta_A,B_0+\delta_Bg^{-1})\).
        On the blocks of the perturbation this is \((a_{21},a_{22})\mapsto(ha_{21},ha_{22})\) and \((b_{12},b_{22})\mapsto(b_{12}h^{-1},b_{22}h^{-1})\), and it fixes
        \(a_{11},a_{12},b_{11},b_{21}\). This preserves the three equations defining \(\gaugeSlice\).
    \end{proof}

By Lemma \ref{lem:slice-linearization}.(1), the linear part of \(\activeCoord\) on \(\gaugeSlice\) comes from \(\gaugeSlice_M\) alone, and it has rank \(s\) as in \Cref{thm:reduction}.
The variables \((a_{22},b_{22})\) do not enter to first order, but only through the quadratic term \(\delta_B\delta_A\) of \(BA\).
To see how they enter \(\activeCoord\), we compute \(BA-B_0A_0\) on \(\gaugeSlice\) exactly.

\begin{cor}\label{cor:residual-block}
    On \(\gaugeSlice\) near \((A_0,B_0)\), set \(u_{11}\coloneqq2a_{11}+a_{11}^2\), \(u_{12}\coloneqq(I_{\rho'}+a_{11})a_{12}\) and \(u_{21}\coloneqq b_{21}(I_{\rho'}+a_{11})\). Then
    \[
        BA-B_0A_0=\begin{pmatrix}u_{11}&u_{12}\\u_{21}&u_{21}(I_{\rho'}+u_{11})^{-1}u_{12}+b_{22}a_{22}\end{pmatrix},
    \]
    and \((a_{11},a_{12},b_{21},a_{22},b_{22})\mapsto(u_{11},u_{12},u_{21},a_{22},b_{22})\) is a local real-analytic diffeomorphism at \(0\).
    In particular, on \((A_0,B_0)+\gaugeSlice_{\mathrm{res}}\), \[
        BA - B_0A_0 = \begin{pmatrix}
            0 & 0 \\
            0 & b_{22}a_{22}
        \end{pmatrix}.
    \] Moreover, the image of the residual map \[
        (a_{22},b_{22})\mapsto b_{22}a_{22} \in \RR^{(m-\rho')\times(d-\rho')}
    \] is exactly the determinantal variety of matrices of rank at most \(r-\rho'\), and the map is invariant under
    the gauge action of \(\GenLinGrp{r-\rho'}\) on \((a_{22},b_{22})\).
\end{cor}

    \begin{proof}
        On \(\gaugeSlice\), we may write \(A=\begin{psmallmatrix}I_{\rho'}+a_{11}&a_{12}\\0&a_{22}\end{psmallmatrix}\) and \(B=\begin{psmallmatrix}I_{\rho'}+a_{11}&0\\b_{21}&b_{22}\end{psmallmatrix}\),
        so \[
            BA-B_0A_0=\begin{pmatrix}(I_{\rho'}+a_{11})^2-I_{\rho'}&(I_{\rho'}+a_{11})a_{12}\\b_{21}(I_{\rho'}+a_{11})&b_{21}a_{12}+b_{22}a_{22}\end{pmatrix}.
        \] Since \((I_{\rho'}+a_{11})^2=I_{\rho'}+u_{11}\), 
        we get \(b_{21}a_{12}=u_{21}(I_{\rho'}+u_{11})^{-1}u_{12}\). 
        The differential of the map \(\Psi\colon(a_{11},a_{12},b_{21},a_{22},b_{22})\mapsto(u_{11},u_{12},u_{21},a_{22},b_{22})\) 
        at \(0\) is \[
            \differential\Psi_0(a_{11},a_{12},b_{21},a_{22},b_{22})=(2a_{11},a_{12},b_{21},a_{22},b_{22}),
        \] which is invertible, so \(\Psi\) is a local real-analytic
        diffeomorphism at \(0\) by the inverse function theorem. 
        On \((A_0,B_0)+\gaugeSlice_{\mathrm{res}}\) we have \(a_{11}=a_{12}=b_{21}=0\), so we get \[
            BA-B_0A_0=\begin{psmallmatrix}0&0\\0&b_{22}a_{22}\end{psmallmatrix}.
        \]

        Since \(b_{22}\) has \(r-\rho'\) columns and \(a_{22}\) has \(r-\rho'\) rows, the product \(b_{22}a_{22}\) has rank at most \(r-\rho'\).
        Conversely, a matrix of rank \(k\leq r-\rho'\) in \(\RR^{(m-\rho')\times(d-\rho')}\) factors as \(UV\) 
        with \(U\in\RR^{(m-\rho')\times k}\) and \(V\in\RR^{k\times(d-\rho')}\), and padding \(U\) and \(V\) with zeros 
        yields a pair \((b_{22},a_{22})\) with \(b_{22}a_{22}=UV\). So the image of \((a_{22},b_{22})\mapsto b_{22}a_{22}\)
        is exactly the set of matrices of rank at most \(r-\rho'\).
        Finally, \(b_{22}a_{22}\) is invariant under the action in Lemma \ref{lem:slice-linearization}.(2).
    \end{proof}

Together, Lemma \ref{lem:local-gauge-slice} and Corollary \ref{cor:residual-block} reduce the local threshold at an equal-rank point to the threshold of a single function on the slice.
In the normal form above, this function is
\begin{equation}\label{eq:nested-residual}
    \norm{\Pi_+\vector{\begin{pmatrix}u_{11}&u_{12}\\u_{21}&u_{21}(I_{\rho'}+u_{11})^{-1}u_{12}+b_{22}a_{22}\end{pmatrix}}}^2,
\end{equation}
where \(u_{11},u_{12},u_{21},a_{22},b_{22}\) are the variables of Corollary \ref{cor:residual-block}.

By \Cref{thm:reduction}, we know the threshold of \eqref{eq:nested-residual} up to 
the threshold \((\lambda_\chi,\multiplicity_\chi)\) of the residual \(\chi\).
It is trivial for \(\rho'=r\) and \(\kappa=0\), which is Corollary \ref{cor:clean-case}, and under the hypothesis of Proposition \ref{prop:aw-transfer} the threshold of the model is known without it.
If \(\ker\Pi_+\subseteq\image{M}\), the threshold of \eqref{eq:nested-residual} is that of an explicit function by Proposition \ref{prop:exact-residual}, and in general it is bounded by Proposition \ref{prop:kappa-bound}.

\begin{prop}\label{prop:exact-residual}
    In the normal form above, suppose that every element of \(\ker\Pi_+\) has vanishing lower right block, that is, \(\ker\Pi_+\subseteq\image M\).
    Let
    \[
        f(z,a_{22},b_{22})=\frobnorm{z_{21}(I_{\rho'}+z_{11})^{-1}z_{12}+b_{22}a_{22}}^2,
    \]
    where \(z=\begin{psmallmatrix}z_{11}&z_{12}\\z_{21}&0\end{psmallmatrix}\in\ker\Pi_+\).
    Let \((\lambda_0,\multiplicity_0)=\rlct_0(f)\), or \((\lambda_0,\multiplicity_0)=(0,1)\) if \(f\) vanishes identically.
    Then \(\kappa=\dim\ker\Pi_+\) and
    \[
        \rlct_{(A_0,B_0)}(\activeCoord)=\Bigl(\frac{\rho'(m+d-\rho')-\kappa}2+\lambda_0,\multiplicity_0\Bigr).
    \]
\end{prop}

    \begin{proof}
        Choose a complement \(Q\) of \(\ker\Pi_+\) in \(\image M\), and let \(N\) be the space of matrices supported on the lower right block.
        Then \(\RR^{m\times d}=Q\oplus\ker\Pi_+\oplus N\).
        As \(\Pi_+\) vanishes on \(\ker\Pi_+\) and is injective on \(Q\oplus N\), we have
        \[
            \norm{\Pi_+\vector Z}\asymp\norm{\pi_QZ}+\norm{\pi_NZ}
        \]
        for the projections along this decomposition.
        By Lemma \ref{lem:local-gauge-slice}.(3), we may compute \(\rlct_{(A_0,B_0)}(\activeCoord)\) on the slice.
        In the coordinates of Corollary \ref{cor:residual-block}, write the upper and left blocks \((u_{11},u_{12},u_{21})\) as \(u'+z\) with \(u'\in Q\) and \(z\in\ker\Pi_+\).
        With \(h=u_{21}(I_{\rho'}+u_{11})^{-1}u_{12}+b_{22}a_{22}\), this gives
        \[
            \norm{\activeCoord}^2\asymp\norm{u'}^2+\norm{h(u',z,a_{22},b_{22})}^2.
        \]
        By the analytic Hadamard lemma, \(h(u',z,a_{22},b_{22})-h(0,z,a_{22},b_{22})\) is bounded by a multiple of \(\norm{u'}\), so
        \[
            \norm{\activeCoord}^2\asymp\norm{u'}^2+\norm{h(0,z,a_{22},b_{22})}^2.
        \]
        Since \(\ker\Pi_+\subseteq\image M\), we have \(\kappa=\dim\ker\Pi_+\), and \(\Pi_+M\) has rank \(\dim Q\).
        So \(u'\) has \(s=\rho'(m+d-\rho')-\kappa\) coordinates by \Cref{thm:reduction}, and they are disjoint from the variables \((z,a_{22},b_{22})\).
        Now Lemma \ref{lem:sum-product-rule} proves the claim.
    \end{proof}

If \(\ker\Pi_+\not\subseteq\image M\), then some element of \(\ker\Pi_+\) has a nonzero lower right block, and the decomposition used for Proposition \ref{prop:exact-residual} is not available.
In those cases, we only bound the local threshold.

The lower bound follows from \Cref{thm:reduction} when \(\lambda_\chi = 0\).
For the upper bound, we compare with Proposition \ref{prop:rlct_aw} for \((m,d,r,\rho')\), 
in which each of the \(\rho'(m+d-\rho')\) directions in \(\image M\) contribute \(\frac12\).
The \(\kappa\) directions in \(\image M\cap\ker\Pi_+\) enter \(\activeCoord\) only at second order, 
so each of them contribute \(\frac14\) instead.
By Lemma \ref{lem:sum-product-rule} contributions add, resulting in the following bounds.

\begin{prop}\label{prop:kappa-bound}
    In the normal form above, let \((\lambda,\multiplicity)=\rlct_{(A_0,B_0)}(\activeCoord)\).
    Let \((\lambda',\multiplicity')\) be the RLCT of Proposition \ref{prop:rlct_aw} for \((m,d,r,\rho')\).
    Then
    \[
        \frac{\rho'(m+d-\rho')-\kappa}2\leq\lambda\leq\lambda'-\frac{\kappa}4.
    \]
\end{prop}

    \begin{proof}
        The first inequality of the proposition is \Cref{thm:reduction}, as \(\lambda_\chi\geq0\).
        For the second, we compute \((\lambda,\multiplicity)\) on the slice by Lemma \ref{lem:local-gauge-slice}.(3),
        in the coordinates of Corollary \ref{cor:residual-block}.

        Write \(\iota(u_{11},u_{12},u_{21})\) for the matrix with blocks \(u_{11},u_{12},u_{21}\) and lower right block \(0\),
        so that \(\image M\) consists of these matrices and \(\iota\) is an isometry.
        Choose orthogonal linear coordinates \((u',u'')\) on the space of \((u_{11},u_{12},u_{21})\)
        such that \(\iota(u'')\) ranges over \(\image M\cap\ker\Pi_+\).
        Then \(u''\) has \(\kappa\) entries, and \(u'\) has the other \(\rho'(m+d-\rho')-\kappa\).

        For \(u=(u_{11},u_{12},u_{21})\) with \(\norm{u}\leq\frac12\) we have \(\norm{(I_{\rho'}+u_{11})^{-1}}\leq2\), hence
        \[
            \frobnorm{u_{21}(I_{\rho'}+u_{11})^{-1}u_{12}}\leq\norm{u}^2=\norm{u'}^2+\norm{u''}^2.
        \]
        By Corollary \ref{cor:residual-block} and \(\Pi_+\iota(u'')=0\), we have near \(0\)
        \begin{align*}
            \norm{\activeCoord}&\leq\norm{u'}+\norm{u'}^2+\norm{u''}^2+\frobnorm{b_{22}a_{22}}\\
            &\leq2\norm{u'}+\norm{u''}^2+\frobnorm{b_{22}a_{22}},
        \end{align*}
        so
        \[
            \norm{\activeCoord}^2\leq12\bigl(\norm{u'}^2+\norm{u''}^4+\frobnorm{b_{22}a_{22}}^2\bigr).
        \]
        By \cite[Prop.~3.4]{MR3614481}, \((\lambda,\multiplicity)\) is at most the threshold of the right-hand side.

        Let \((\lambda'',\multiplicity'')\) be the RLCT of Proposition \ref{prop:rlct_aw} for \((m-\rho',d-\rho',r-\rho',0)\).
        The function \(\frobnorm{b_{22}a_{22}}^2\) is homogeneous of degree \(4\) in \((a_{22},b_{22})\),
        so its local threshold is the same at \(x\) and \(tx\) for \(t>0\) \cite[Prop.~3.8 and Lemma~2.4]{MR3614481}.
        By \cite[Prop.~2.5]{MR3614481}, its threshold on a closed ball around \(0\) is therefore its local threshold at \(0\),
        which is \((\lambda'',\multiplicity'')\) by Proposition \ref{prop:rlct_aw}.

        In \eqref{eq:aoyagi-watanabe}, the range of \(s\) is the same for both quadruples,
        and the two bracketed expressions differ by \((m+d)\rho'-\rho'^2=\rho'(m+d-\rho')\),
        so \(\lambda'=\frac{\rho'(m+d-\rho')}2+\lambda''\).
        By Lemma \ref{lem:sum-product-rule} for the variables \(u'\), \(u''\) and \((a_{22},b_{22})\), we get
        \[
            \lambda\leq\frac{\rho'(m+d-\rho')-\kappa}2+\frac{\kappa}4+\lambda''=\lambda'-\frac{\kappa}4. \qedhere
        \]
    \end{proof}

\begin{example}
    In Example \ref{ex:two-by-two} we have \(\rho'=1\), so \(\lambda'=\frac32\) by \Cref{prop:rlct_aw}.
    For \(\kappa=1\), Proposition \ref{prop:kappa-bound} gives \(1\leq\lambda\leq\frac54\), and \(\lambda=1\) by \Cref{tab:two-by-two} attains the lower bound.
    For \(\kappa=2\), it gives \(\frac12\leq\lambda\leq1\), and \(\lambda=1\) attains the upper bound.
    In this example \(\ker\Pi_+\subseteq\image M\), so these values also follow from Proposition \ref{prop:exact-residual}.
\end{example}

\section{Global Geometry and Determinantal Sections}
\label{sec:global}
\Cref{sec:local} studies the local RLCT of \(\activeCoord\) at a single equal-rank point of \(W_\Pi\).
The RLCT of the model on a parameter domain is the minimum of the local RLCTs at the points of \(W_\Pi\) in that domain,
so we now pass from one point to all of \(W_\Pi\).

Two properties of \(W_\Pi\) make this possible.
First, \(W_\Pi\) is the preimage of the linear section \(\linearSection\) under \((A,B)\mapsto BA\), and \(\activeCoord\) depends on \((A,B)\) only through \(BA\).
Second, the gauge orbit of every point of \(W_\Pi\) comes arbitrarily close to an equal-rank point with the same product.
Together, these reduce the global RLCT to the local RLCTs at equal-rank points over the products in \(\linearSection\).

The products in \(\linearSection\) need not have the rank of \(B_0A_0\), and \(\linearSection\) can be singular away from \(B_0A_0\).
The minimum is therefore taken over all of \(\linearSection\), not only at the true update.

\subsection{Reduction to Equal-Rank Points}
\Cref{thm:reduction} concerns equal-rank points, whereas the global threshold is a minimum over all points of \(W_\Pi\).
A point of \(W_\Pi\) need not be of this kind, for instance when \(\image A\cap\ker B\neq0\).
The images of such a point under a one-parameter subgroup of \(\GenLinGrp{r}\) converge to an equal-rank point with the same product, whose local threshold is at most that of the original point.
So it suffices to consider equal-rank points.

\begin{lemma}\label{lem:degeneration}
    Let \((A,B)\in\RR^{r\times d}\times\RR^{m\times r}\). There is a one-parameter subgroup \(t\mapsto g(t)\) of \(\GenLinGrp{r}\) such that
    \(g(t)\cdot(A,B)\) converges for \(t\to0\) to an equal-rank point \((A^*,B^*)\) with \(B^*A^*=BA\).
\end{lemma}

    \begin{proof}
        Let \(E_2\coloneqq\image A\cap\ker B\), and choose complements \(\image A=E_1\oplus E_2\) and \(\ker B=E_2\oplus E_3\). The sum
        \(E_1\oplus E_2\oplus E_3\) is direct, since \(E_3\cap\image A\subseteq E_3\cap E_2=0\). Choose a complement \(E_4\) of it in \(\RR^r\) and let
        \(\pi_i\) be the corresponding projections. 
        
        For \(t>0\), let \(g(t)\coloneqq\pi_1+t\pi_2+\pi_3+t^{-1}\pi_4\). 
        Since the \(\pi_i\) are complementary projections, this satisfies \(g(st)=g(s)g(t)\) and \(g(1)=I_r\),
        so \(t\mapsto g(t)\) is a one-parameter subgroup of \(\GenLinGrp{r}\) with \(g(t)^{-1}=g(t^{-1})\).
        Since \(\image A\subseteq E_1\oplus E_2\), we have \(g(t)A=\pi_1A+t\pi_2A\), and since \(B\) vanishes on \(E_2\oplus E_3\), we have \(Bg(t)^{-1}=B\pi_1+tB\pi_4\).
        These are the components of \(g(t)\cdot(A,B)\), and as \(t\to0\) they converge to \(A^*\coloneqq\pi_1A\) and \(B^*\coloneqq B\pi_1\).

        As a result, the product is unchanged, \(B^*A^*=BA\) since \(B\) vanishes on \(E_2\) 
        and \(\image A\subseteq E_1\oplus E_2\).
        The image of \(A^*\) is \(E_1\), and \(B\) is injective on \(E_1\) as \(E_1\cap\ker B\subseteq E_1\cap E_2=0\).
        
        Hence \(A^*\), \(B^*\) and \(BA\) have the images \(E_1\), \(B(E_1)\) and \(B(E_1)\), 
        so all three have rank \(\dim E_1\), and \((A^*,B^*)\) is an equal-rank point.
    \end{proof}

The gauge action preserves the ranks of \(A\), \(B\) and \(BA\), so the orbit of \((A,B)\) contains an equal-rank point only if \((A,B)\) is one.
The equal-rank point of Lemma \ref{lem:degeneration} is therefore reached only in the limit, where the local threshold may drop, resulting in the following upper bound.

\begin{cor}\label{cor:degeneration}
    For \((A,B)\in W_\Pi\) and \((A^*,B^*)\) as in Lemma \ref{lem:degeneration}, we have \((A^*,B^*)\in W_\Pi\) and
    \(\rlct_{(A^*,B^*)}(\activeCoord)\leq\rlct_{(A,B)}(\activeCoord)\).
\end{cor}

    \begin{proof}
        Both \(W_\Pi\) and \(\activeCoord\) only depend on \(BA\), so \((A^*,B^*)\in W_\Pi\) as \(B^*A^*=BA\).
        The gauge action of \(g\in\GenLinGrp{r}\) is linear and invertible and preserves \(\activeCoord\), 
        so the local RLCT of \(\activeCoord\) is the same at all points 
        \(g(t)\cdot(A,B)\) \cite[Prop.~3.8 and Lemma~2.4]{MR3614481}.
        
        By the definition of local thresholds and \cite[Prop.~2.5]{MR3614481}, the local threshold of \(\activeCoord\) at \((A^*,B^*)\)
        is at most its local threshold at every interior point of a sufficiently small compact neighborhood of \((A^*,B^*)\).
        For small \(t\), the point \(g(t)\cdot(A,B)\) is such a point, and its local threshold is that of \((A,B)\) by the above.
        Hence the local threshold at \((A^*,B^*)\) is at most that at \((A,B)\).
    \end{proof}

The equal-rank point in Lemma \ref{lem:degeneration} depends on the choice of the complements \(E_1\), \(E_3\) and \(E_4\).
All choices give the same product \(BA\), and the next lemma shows that an equal-rank point is determined by its product up to the gauge action.

\begin{lemma}\label{lem:single-orbit}
    Let \(Y\in\RR^{m\times d}\) with \(\rho'\coloneqq\rank Y\leq r\). The equal-rank points \((A,B)\in\RR^{r\times d}\times\RR^{m\times r}\) with \(BA=Y\) form a single gauge orbit.
\end{lemma}

    \begin{proof}
        Since \(\rank BA=\rank A\), we have \(\image A\cap\ker B=0\), and \(\dim\image A+\dim\ker B=r\), so
        \(\RR^r=\image A\oplus\ker B\). The row space of \(Y=BA\) is contained in the row space of \(A\), and both have dimension \(\rho'\), so they agree.

        Choose \(g\in\GenLinGrp{r}\) with \(g(\image A)=\RR^{\rho'}\times0\) and \(g(\ker B)=0\times\RR^{r-\rho'}\). Then
        \[
            gA=\begin{pmatrix}A'\\0\end{pmatrix},\qquad Bg^{-1}=\begin{pmatrix}B'&0\end{pmatrix}
        \]
        with \(A'\in\RR^{\rho'\times d}\) of rank \(\rho'\), \(B'\in\RR^{m\times\rho'}\) and \(B'A'=Y\). The row space of \(A'\) is that of \(Y\).

        Fix \(A_0'\in\RR^{\rho'\times d}\) whose rows form a basis of the row space of \(Y\), and let \(B_0'\) be the unique matrix with \(B_0'A_0'=Y\). Then
        \(A'=hA_0'\) for a unique \(h\in\GenLinGrp{\rho'}\). From \(B'hA_0'=Y=B_0'A_0'\) and since \(A_0'\) has a right inverse, \(B'h=B_0'\). Hence
        \(\diag(h^{-1},I_{r-\rho'})\,g\) maps \((A,B)\) to \(\bigl(\begin{psmallmatrix}A_0'\\0\end{psmallmatrix},\begin{pmatrix}B_0'&0\end{pmatrix}\bigr)\),
        which does not depend on \((A,B)\).
        So any two such points are mapped to the same pair by elements of \(\GenLinGrp{r}\), and hence lie in one gauge orbit.
        Conversely, the gauge action preserves \(BA\) and the ranks of \(A\) and \(B\), so every pair in this orbit is again an equal-rank point with product \(Y\).
    \end{proof}

Since \(\activeCoord\) depends on \((A,B)\) only through \(BA\), it is preserved by the gauge action, 
so its local threshold is constant along a gauge orbit.
As a result, the local threshold at an equal-rank point does not depend on which equal-rank point over \(Y\) we take, 
in particular not on the choice of complements in Lemma \ref{lem:degeneration}.
\begin{cor}\label{cor:orbit-invariance}
    For any two equal-rank points \(X',X''\in W_\Pi\) with the same product, the local thresholds of \(\activeCoord\) agree.
\end{cor}

    \begin{proof}
        By Lemma \ref{lem:single-orbit}, \(X''=g\cdot X'\) for some \(g\in\GenLinGrp{r}\). As in the proof of Corollary \ref{cor:degeneration}, the gauge action of \(g\) is linear, invertible and
        preserves \(\activeCoord\), so the local thresholds at \(X'\) and \(X''\) agree \cite[Prop.~3.8 and Lemma~2.4]{MR3614481}.
    \end{proof}

Together, Corollary \ref{cor:degeneration} and Corollary \ref{cor:orbit-invariance} show that the global threshold is determined by the products \(Y\) in
the linear section \(\linearSection\): every point of \(W_\Pi\) over \(Y\) degenerates to an equal-rank point over \(Y\) with no larger
threshold, and all equal-rank points over \(Y\) have the same threshold.

\subsection{The Global Threshold}
By Corollary \ref{cor:degeneration} and Corollary \ref{cor:orbit-invariance}, the RLCT of the model is determined by the local thresholds at equal-rank points, as follows.

\begin{theorem}\label{thm:global-rlct}
    Let \(\Omega\) be compact with \(W_\Pi\cap\Omega\neq\emptyset\).
    \begin{itemize}
        \item[(a)] We have \(\rlct_\Omega(K)\leq\rlct_{X'}(\activeCoord)\) for every equal-rank point 
            \(X'=(A',B')\in W_\Pi\) in the interior of \(\Omega\).
        \item[(b)] There exists \((A,B)\in W_\Pi\cap\Omega\) such that \(\rlct_{X'}(\activeCoord)\leq\rlct_\Omega(K)\) 
            for every equal-rank point \(X'=(A',B')\) with \(B'A'=BA\).
        \item[(c)] If one of the points \(X'\) in (b) lies in the interior of \(\Omega\), then \(\rlct_\Omega(K)=\rlct_{X'}(\activeCoord)\), and this is the minimum of the values in (a).
    \end{itemize}
\end{theorem}

    \begin{proof}
        By Lemma \ref{lem:sandwich} and \cite[Cor.~3.5]{MR3614481} we have \(\rlct_\Omega(K)=\rlct_\Omega(\activeCoord)\), so we may replace \(K\) by \(\activeCoord\) in (a) to (c).
        \begin{itemize}
            \item[(a)] The threshold of \(\activeCoord\) at \(X'\) relative to \(\Omega\) is \(\rlct_{X'}(\activeCoord)\), as \(X'\) is interior.
                Hence \(\rlct_\Omega(\activeCoord)\leq\rlct_{X'}(\activeCoord)\) by \cite[Prop.~2.5]{MR3614481}.
            \item[(b)] By \cite[Prop.~2.5]{MR3614481} there is \(q\in W_\Pi\cap\Omega\) at which the threshold of \(\activeCoord\) relative to \(\Omega\) equals \(\rlct_\Omega(\activeCoord)\).
                Write \(q=(A,B)\) and let \(q^*=(A^*,B^*)\) be the point obtained from \(q\) by Lemma \ref{lem:degeneration}.
                By this lemma, \(q^*\) is an equal-rank point with \(B^*A^*=BA\).
                Then \(\rlct_{q^*}(\activeCoord)\leq\rlct_q(\activeCoord)\) by Corollary \ref{cor:degeneration},
                and \(\rlct_q(\activeCoord)\leq\rlct_\Omega(\activeCoord)\) by \cite[Prop.~3.3]{MR3614481}.
                Since \(q^*\) and every \(X'\) as in (b) are equal-rank points with product \(BA\),
                they have the same threshold by Corollary \ref{cor:orbit-invariance}.
            \item[(c)] This follows from (a) and (b). \qedhere
        \end{itemize}
    \end{proof}

The next example has \(\kappa=0\) and \(\rho<r\). The residual function still contains terms of higher order than its leading term, and these do not change the
local threshold. The example also shows that the local threshold at the true update need not give the threshold of the model.

\begin{example}\label{ex:three-by-three}
    Let \(d=m=3\), \(r=2\) and \(\rho = 1\). Assume that \(A_0 = \begin{psmallmatrix}
        1 & 0 & 0 \\
        0 & 0 & 0
    \end{psmallmatrix}\) and \(B_0 = \begin{psmallmatrix}
        1 & 0 \\
        0 & 0 \\
        0 & 0
    \end{psmallmatrix}\). For the linearization \(M\) we obtain the following expression \[
        M\delta = \begin{pmatrix}
            a_{11} + b_{11} & a_{12} & a_{13} \\
                  b_{21}    &   0    &   0    \\
                  b_{31}    &   0    &   0
        \end{pmatrix}.
    \] In other words, \(\image{M} = \langle\, E_{11},E_{12},E_{13},E_{21},E_{31} \,\rangle\). For \(\ker\Pi_+\)
    we pick the following three-dimensional linear space \[
        L = \langle\,
            \ell_1 \coloneqq E_{11} + E_{22},
            \ell_2 \coloneqq E_{13} + E_{23},
            \ell_3 \coloneqq E_{31} + E_{32}
        \,\rangle.
    \] The \(\ell_i\) are pairwise orthogonal due to disjoint support and \(\norm{\ell_i}^2 = 2\), so
    \[
        \Pi_+(X) = X - \frac12\sum_{i=1}^3\langle X,\ell_i\rangle\ell_i.
    \] In particular, \(L\cap\image{M} = 0\), so \(\Pi_+\) is injective when restricted to \(\image{M}\).
    For \(D \coloneqq X - B_0A_0\), the six independent coordinates of \(\Pi_+(D)\) are \[
        \activeCoord_1 = \frac12(D_{11}-D_{22}),\activeCoord_2 = \frac12(D_{13}-D_{23}),
        \activeCoord_3 = \frac12(D_{31}-D_{32}),\activeCoord_4 = D_{12},\activeCoord_5 = D_{21},\activeCoord_6 = D_{33}.
    \] Thus, the KL divergence can be written as \[
        K = \frac12\norm{\Pi_+(D)}^2 
          = \activeCoord_1^2 + \activeCoord_2^2 + \activeCoord_3^2 
          + \frac12\activeCoord_4^2 + \frac12\activeCoord_5^2 + \frac12\activeCoord_6^2.
    \]

    Using Lemma \ref{lem:local-gauge-slice}, we work locally subject to \[
        a_{21}=0,\qquad b_{12}=0,\qquad b_{11}=a_{11}.
    \] With this, a nearby point of \((A_0,B_0)\) has the form \[
        A = \begin{pmatrix}
            1+a_{11} & a_{12} & a_{13}\\
            0 & a_{22} & a_{23}
        \end{pmatrix},
        \qquad
        B = \begin{pmatrix}
            1+a_{11} & 0\\
            b_{21} & b_{22}\\
            b_{31} & b_{32}
        \end{pmatrix}.
    \] 
    Computing \(D = BA - B_0A_0\) and substituting into the active coordinates \(\activeCoord_i\) gives 
    \begin{align*}
        \activeCoord_1
        &=\frac12\left(
            2a_{11}+a_{11}^2-b_{21}a_{12}-b_{22}a_{22}
          \right), &
        \activeCoord_2
        &=\frac12\left(
            (1+a_{11})a_{13}-b_{21}a_{13}-b_{22}a_{23}
          \right),\\
        \activeCoord_3
        &=\frac12\left(
            (1+a_{11})b_{31}-b_{31}a_{12}-b_{32}a_{22}
          \right), &
        \activeCoord_4&=(1+a_{11})a_{12},\\
        \activeCoord_5&=(1+a_{11})b_{21}, &
        \activeCoord_6&=b_{31}a_{13}+b_{32}a_{23}.
    \end{align*} Locally, we can eliminate the coordinates of \(\gaugeSlice_M\) using those of 
    \(\gaugeSlice_{\mathrm{res}}\) since the Jacobian of the coordinate change \[
        (a_{11},a_{12},a_{13},b_{21},b_{31},a_{22},a_{23},b_{22},b_{32})
        \mapsto
        (\activeCoord_1,\activeCoord_2,\activeCoord_3,\activeCoord_4,\activeCoord_5,a_{22},a_{23},b_{22},b_{32})
    \] is invertible. 

    From \(\activeCoord_4 = \activeCoord_5 = 0\), and since \(1+a_{11}\) is non-zero around the origin, we obtain
    that \(a_{12} = b_{21} = 0\). Then \(\activeCoord_1=0\) gives \[
        a_{11}^2 + 2a_{11} = a_{22}b_{22}.
    \] Choosing the branch \(a_{11} = \sqrt{1 + a_{22}b_{22}} - 1\) where all slice coordinates vanish then gives \[
        a_{13} = \frac{b_{22}a_{23}}{\sqrt{1 + a_{22}b_{22}}}
        \quad\text{and}\quad
        b_{31} = \frac{b_{32}a_{22}}{\sqrt{1 + a_{22}b_{22}}}.
    \] Substituting this into the expression for \(\activeCoord_6\) then gives \[
        \activeCoord_6 = b_{32}a_{23}\frac{1 + 2a_{22}b_{22}}{1+a_{22}b_{22}}
    \] with series expansion \[
        \activeCoord_6 = b_{32}a_{23} 
                + a_{22}a_{23}b_{22}b_{32} 
                + \mathcal{O}\left(\norm{(a_{22},a_{23},b_{22},b_{32})}^6\right).
    \] Since the higher-order terms of \(\activeCoord_6\) come from the factor \(\frac{1 + 2a_{22}b_{22}}{1+a_{22}b_{22}}\),
    which is a unit near the origin, we have \[
        \rlct_0(\activeCoord_6^2) = \rlct_0((b_{32}a_{23})^2) = \left(\frac12,2\right).
    \] Together with the transverse quadratic coordinates \(\activeCoord_1,\dots,\activeCoord_5\), we get \[
        (\lambda,\multiplicity)_{A_0,B_0} = \left(\frac52,1\right) + \left(\frac12,2\right) = (3,2).
    \]

    Note that this is only a computation for the \emph{local} RLCT at a true parameter point \((A_0,B_0)\). For the (global)
    RLCT, we need to inspect the entire linear section \[
        \linearSection = (B_0A_0 + L) \cap \detVar{\leq 2}.
    \] For the affine space \(B_0A_0 + L\), consider a generic matrix \[
        W(s) 
        = B_0A_0 + s_1\ell_1 + s_2\ell_2 + s_3\ell_3
        = \begin{pmatrix}
            1 + s_1 & 0   & s_2 \\
            0       & s_1 & s_2 \\
            s_3     & s_3 & 0
        \end{pmatrix}
    \] which has determinant \(\det{W(s)} = -s_2s_3(1+2s_1)\). Thus, the section \(\linearSection\) is the union of
    the three planes \[
        s_2 = 0,\quad s_3 = 0,\quad s_1 = -\frac12.
    \] At the triple intersection point \(P\) the matrix \(W(s)\) has rank \(2\), and after recentering, the determinant is \(
        \det{W} = -2\delta_1s_2s_3
    \) which means the local residual has a three-fold normal crossing, resulting in \[
        \rlct_0((\delta_1s_2s_3)^2) = \left(\frac12,3\right).
    \] Similar to above, the local RLCT at \(P\) is \(
        (\lambda,\multiplicity)_P = (3,3)
    \) which turns out to be the RLCT for this specific model.
\end{example}

\section{RLCT Estimation and Experiments}
\label{sec:experiments}
We now study the singular geometry developed in the preceding sections numerically, which requires estimating the RLCT from finite-sample posterior computations.  We first review existing single- and multi-temperature RLCT estimators and introduce a novel \emph{curvature-corrected multi-temperature estimator} designed to account for finite-sample variation of the effective learning coefficient across inverse temperatures.  We then use these estimators to study the forward-backward covariance and learning complexity in three settings: synthetic trace regression with frozen neural-network features; transfer learning on rotated digits; and prompt-based surgical fine-tuning of a pretrained transformer.

\subsection{Estimating the RLCT}\label{sec:rlct}

Consider the tempered posterior
\[
\pi_{\beta}(w \mid X^{n}) \propto \exp\bigl(-\beta H_{n}(w)\bigr)\,\pi(w),
\qquad
H_{n}(w):=-\sum_{i=1}^{n}\log p(X_{i}\mid w),
\]
and write $\mathbb{E}^{(\beta)}_{\mathrm{pos}}$ and $\mathbb{V}^{(\beta)}_{\mathrm{pos}}$ for its mean and variance, respectively.
Let $\beta^{\star}=1/\log n$.
All estimators below rely on two relations.
First, by Theorem~4 of \cite{Watanabe:2013}, for $\beta=\Theta(1/\log n)$, we have
\begin{align}
  \mathbb{E}^{(\beta)}_{\mathrm{pos}}[H_{n}(w)]
  = H_{n}(w_{0})+\frac{\lambda}{\beta}+U_{n}\sqrt{\frac{\lambda}{2\beta}}+\mathcal{O}_{P}(1),
  \label{eq:fundamental tempered posterior}
\end{align}
where $w_{0}$ minimizes $\mathbb{E}[H_{n}(w)]$ and $U_{n}$ converges in distribution to the standard normal.
Second, for every $\beta>0$ (Problem~2 in Section~1.9 of \cite{watanabe2018mathematical}), we have
\begin{align}
  \frac{\partial}{\partial (1/\beta)}\mathbb{E}^{(\beta)}_{\mathrm{pos}}[H_{n}(w)]
  = \beta^{2}\,\mathbb{V}^{(\beta)}_{\mathrm{pos}}[H_{n}(w)] =: \lambda(\beta).
  \label{eq:fluctuation-dissipation theorem}
\end{align}
Notice that by \eqref{eq:fundamental tempered posterior}, $\lambda(\beta^{\star})$ approaches $\lambda$ as $n\to\infty$.

\paragraph{Single-Temperature RLCT Estimators.}

We begin by reviewing single-temperature RLCT estimators proposed in the literature.
The \emph{WBIC-type estimator} \citep{lau2025} is given by
\[
\hat{\lambda}_{\mathrm{simple}}
:=\beta^{\star}\bigl\{\mathbb{E}^{(\beta^{\star})}_{\mathrm{pos}}[H_{n}(w)]-H_{n}(\hat{w})\bigr\},
\]
where $\hat{w}$ is a local minimizer of $H_{n}$; it follows from \eqref{eq:fundamental tempered posterior}.
The \emph{heat-capacity estimator} \citep{imai2019estimating,tokuda2022} is
\[
\hat{\lambda}_{\mathrm{heat}}
:=(\beta^{\star})^{2}\,\mathbb{V}^{(\beta^{\star})}_{\mathrm{pos}}[H_{n}(w)]=\lambda(\beta^{\star}),
\]
obtained by formally differentiating \eqref{eq:fundamental tempered posterior} and using \eqref{eq:fluctuation-dissipation theorem}.

\paragraph{Multi-Temperature RLCT Estimators.}

Multi-temperature estimators extend beyond the single-temperature case.  The two-temperature estimator \citep{Watanabe:2013} uses the two temperatures $\beta_{1}=\beta^{\star}$ and $\beta_{2}=c\,\beta^{\star}$ ($c>1$),
and are defined by
\[
\hat{\lambda}_{\mathrm{two}}
:=\frac{\mathbb{E}^{(\beta_{1})}_{\mathrm{pos}}[H_{n}(w)]-\mathbb{E}^{(\beta_{2})}_{\mathrm{pos}}[H_{n}(w)]}
       {\beta_{1}^{-1}-\beta_{2}^{-1}},
\]
which follows from \eqref{eq:fundamental tempered posterior}.
Given a set $\mathcal{B}$ of inverse temperatures around $\beta^{\star}$, the linear multi-temperature (LMT) estimator \citep{wei2022} is the slope
\[
(\hat{a},\hat{\lambda}_{\mathrm{LMT}})
:=\operatorname*{arg\,min}_{a,b}\sum_{\beta\in\mathcal{B}}
\bigl\{\mathbb{E}^{(\beta)}_{\mathrm{pos}}[H_{n}(w)]-a-b\,\beta^{-1}\bigr\}^{2},
\]
which again follows from \eqref{eq:fundamental tempered posterior}.
By \eqref{eq:fluctuation-dissipation theorem}, $\hat{\lambda}_{\mathrm{LMT}}$ is a weighted average of $\lambda(\beta)$ over $\mathcal{B}$.
For finite $n$, however, $\mathbb{E}^{(\beta)}_{\mathrm{pos}}[H_{n}(w)]$ need not be linear in $\beta^{-1}$ over $\mathcal{B}$, i.e., $\lambda(\beta)$ need not be constant there, in particular in deep learning applications, thereby creating the dependence of $\hat{\lambda}_{\mathrm{LMT}}$ on the choice of $\mathcal{B}$.

\paragraph{Curvature-Corrected Multi-Temperature Estimator.}
To mitigate the dependence on $\mathcal{B}$ arising in the linear multi-temperature estimator, we propose the \emph{curvature-corrected multi-temperature (CMT) estimator},
the slope at $\beta^{\star}$ of a local quadratic fit in $\beta^{-1}$:
\[
(\hat{a},\hat{\lambda}_{\mathrm{CMT}},\hat{b}_{2})
:=\operatorname*{arg\,min}_{a,b_{1},b_{2}}\sum_{\beta\in\mathcal{B}}
\bigl\{\mathbb{E}^{(\beta)}_{\mathrm{pos}}[H_{n}(w)]-a-b_{1}(\beta^{-1}-\beta^{\star-1})-b_{2}(\beta^{-1}-\beta^{\star-1})^{2}\bigr\}^{2}.
\]
By \eqref{eq:fluctuation-dissipation theorem}, $\hat{\lambda}_{\mathrm{CMT}}$ estimates $\lambda(\beta^{\star})$, while the quadratic term absorbs the variation of $\lambda(\beta)$ around $\beta^{\star}$.

\subsection{Additional Setup}

In addition to the choice of estimator, we require a prior distribution on the LoRA parameters
$w=(A,B)$ and a sampler for the tempered posteriors.

\paragraph{Choice of Priors and Samplers.}
We consider two priors. The \emph{global} prior $\pi(w)=\mathcal{N}(0,\sigma_{\pi}^{2}I)$ is a global Gaussian prior, and the chains for samplers are started from independent draws. The \emph{localized} prior
$\pi(w)=\mathcal{N}(\hat{w},\gamma^{-1}I)$ is centred at the fitted LoRA point $\hat{w}$ and a sampler starts there,
restricting the posterior to a neighborhood of $\hat{w}$, as in the local learning coefficient of
\cite{lau2025}. 

We use the Metropolis-adjusted Langevin algorithm (MALA, \citet{Roberts1996}) with a Gauss--Newton preconditioner computed at
the initial point.
For comparison, we also run stochastic-gradient
Langevin dynamics (SGLD, \citet{welling2011}) without correction, as in \cite{okuno2023generalization,lau2025}.

The multi-temperature estimators use 
a temperature ladder with size $12$,
$K=12$ inverse temperatures spaced geometrically on
$[\beta^{\star}/10,\,10\beta^{\star}]$, and the fits use the narrow window
$\mathcal{B}=\{\beta:\beta^{\star}/4\le\beta\le4\beta^{\star}\}$. We run $L=4$ independent ladders, either
with replica exchange \citep{swendsen1986,hukushima1996} between neighbouring temperatures or without exchange (independent chains), and
compute every estimator per ladder; the reported value is the mean over ladders and its Monte Carlo
standard error is the standard error across ladders.

\paragraph{Uncertainty Quantification.}

In addition to the Monte Carlo error from independent ladders, we report a model-form error reflecting the assumption of $\lambda(\beta)$ in $\mathcal{B}$:
$\abs{\hat{\lambda}_{\mathrm{CMT}}-\hat{\lambda}_{\mathrm{LMT}}}/2$ for $\hat{\lambda}_{\mathrm{LMT}}$,
and $\abs{\hat{\lambda}_{\mathrm{CMT}}(\mathcal{B})-\hat{\lambda}_{\mathrm{CMT}}(\mathcal{B}')}/2$ for
$\hat{\lambda}_{\mathrm{CMT}}$, where $\mathcal{B}'=\{\beta\in\mathcal{B}:\beta^{\star}/2\le\beta\le2\beta^{\star}\}$
is a narrower window. 

\subsection{Synthetic Trace Regression with Frozen FB Features}

\begin{figure}[h]
    \centering
    \includegraphics[width=0.85\linewidth]{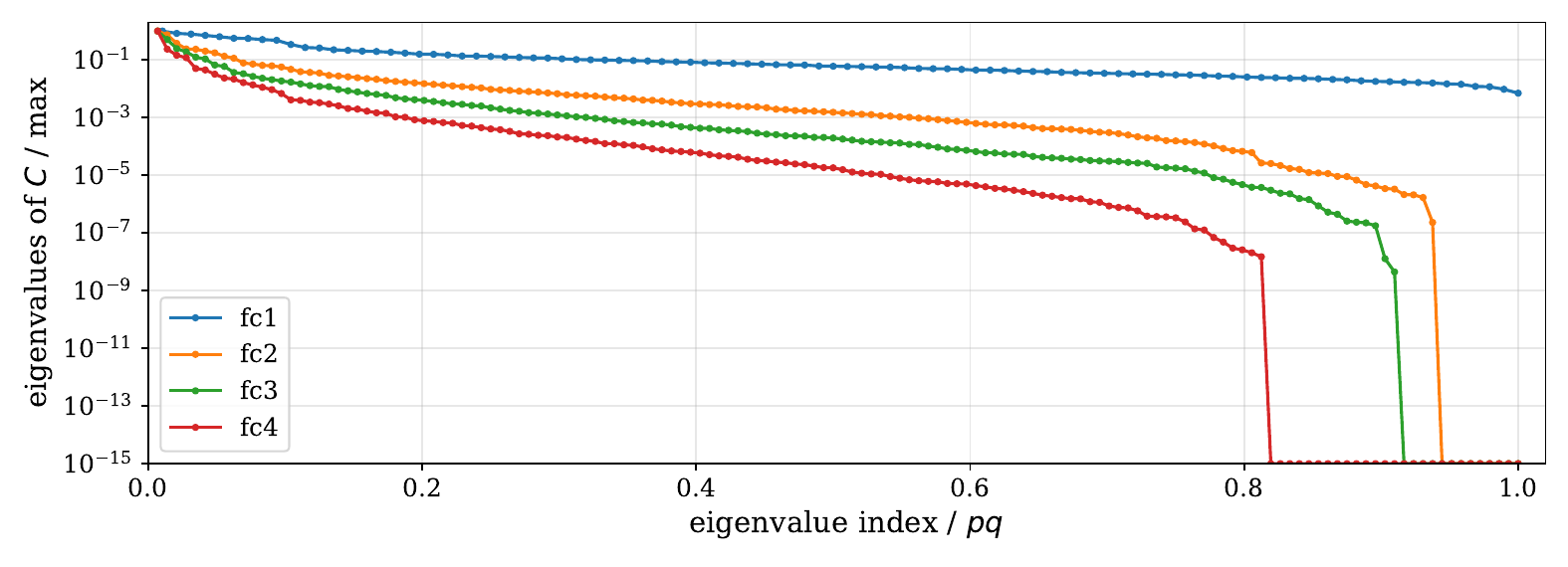}
    \caption{The FB covariance spectra in the synthetic trace regression.}
    \label{fig:fb_spectra_tracereg_surgicalmlp}
\end{figure}

\begin{figure}[h]
    \centering
    \includegraphics[width=0.85\linewidth]{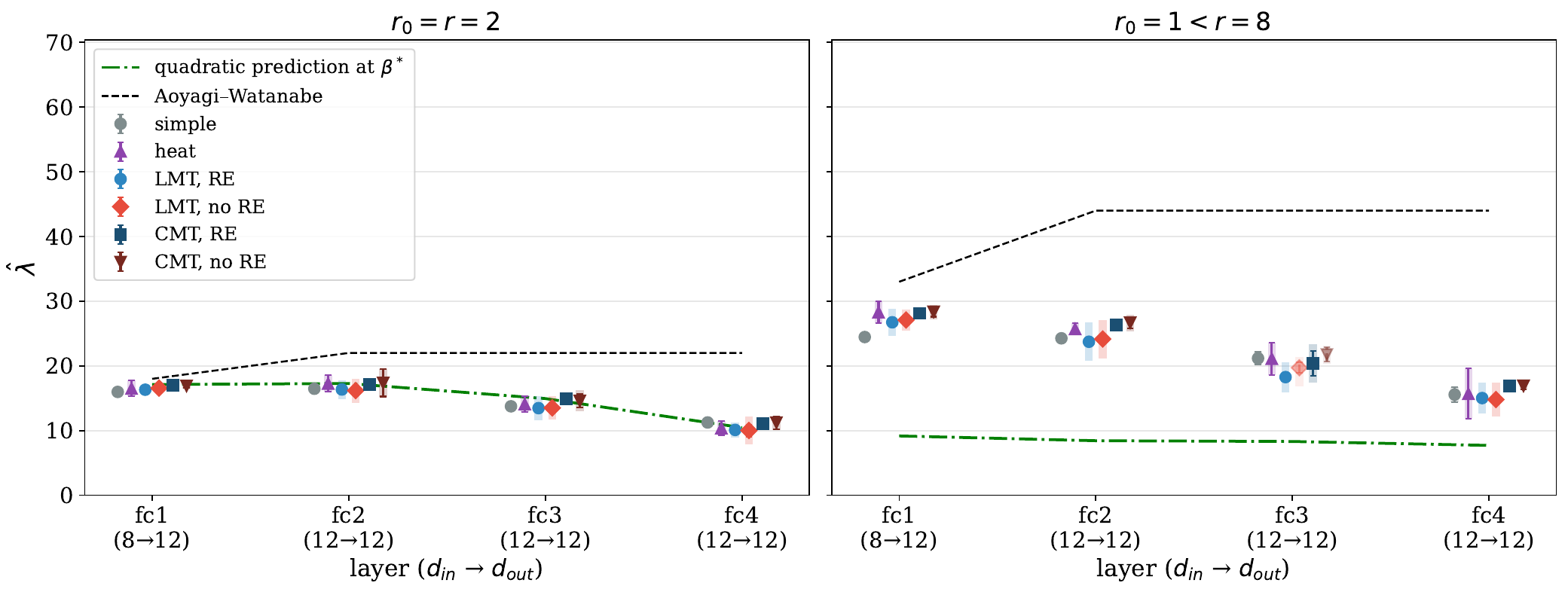}
    \caption{The RLCT estimates based on the localized prior and the MALA in the synthetic trace regression. The left panel shows the results for the equal rank case in which the true rank $r_{0}$ is equal to the LoRA rank $r$. The right panel shows the results for the excess rank case in which $r_{0}<r$.
    For RLCT estimates, we use the methods in \Cref{sec:rlct}. The solid bar is $\pm 2$ Monte Carlo SE. 
    For multi-temperature estimates, the shaded bar describes $\pm 2$(Monte Carlo SE $+$ model-form error).
    The green dashed lines denote the quadratic
prediction $\lambda_{\mathrm{quad}}(\beta^\star)$ of the Laplace approximation, and the black dashed lines denote the \citet{Aoyagi+Watanabe:2005} theoretical values.
    }
    \label{fig:tracereg_rlct_mlp_localized}
\end{figure}

We begin with a synthetic trace regression with the FB features from the structure of a frozen ReLU Multi Layer Perceptron (MLP)
with widths $(8,12,12,12,12,1)$. Its weights are drawn from $\mathcal N(0,2/d_{\mathrm{in}})$, its biases from
$\mathcal N(0,0.25)$, and its inputs of sample size $n=2{,}000$ are drawn from $\mathcal N(0,I_8)$. 
For the $k$-th hidden layer $(k=1,2,3,4)$, 
we compute the input activation, and the backpropagated sensitivity. 
Then, the responses are created to follow the trace regression with this FB feature,
\[
  y_i=\frobprod{g^{k}(x_{i}) h^{k-1}(x_{i})^{\transpose}, B_0A_0}+\varepsilon_i,\qquad \varepsilon_i\sim\mathcal N(0,1),
\]
with a random true update $B_0A_0$ of rank $r_0$. 
We consider two
settings: an equal LoRA rank $r_0=r=2$, and an excess LoRA rank $r_0=1<r=8$. 

Figure~\ref{fig:tracereg_rlct_mlp_localized} shows the RLCT estimates based on the localized prior and the MALA
together with the quadratic
prediction $\lambda_{\mathrm{quad}}(\beta^\star)$ of the Laplace approximation and the \citet{Aoyagi+Watanabe:2005} theoretical value that assumes non-degenerate FB covariance.
For $r_0=r=2$, all estimators agree with $\lambda_{\mathrm{quad}}(\beta^\star)$ in every layer. 
They lie below the value of \citet{Aoyagi+Watanabe:2005}, and the gap grows with layer depth.
For $r_0=1<r=8$, the estimates lie far above
        $\lambda_{\mathrm{quad}}(\beta^\star)$ and below the value of \citet{Aoyagi+Watanabe:2005}.

These results match the two terms of Theorem~\ref{thm:reduction}, $\lambda=s/2+\lambda_\chi$. For $r_0=r$,
the quadratic prediction from the Laplace approximation counts the regular directions at the true point, and its agreement with the
RLCT estimates indicates that the residual term $\lambda_\chi$ is negligible. 
For $r_0<r$, the $r-r_0$ excess LoRA directions enter through the residual determinantal model of
\Cref{cor:residual-block}. In this case, the quadratic prediction misses the contribution $\lambda_\chi$.

\subsection{Transfer to Rotated Digits}

\begin{figure}[h]
    \centering
    \includegraphics[width=0.85\linewidth]{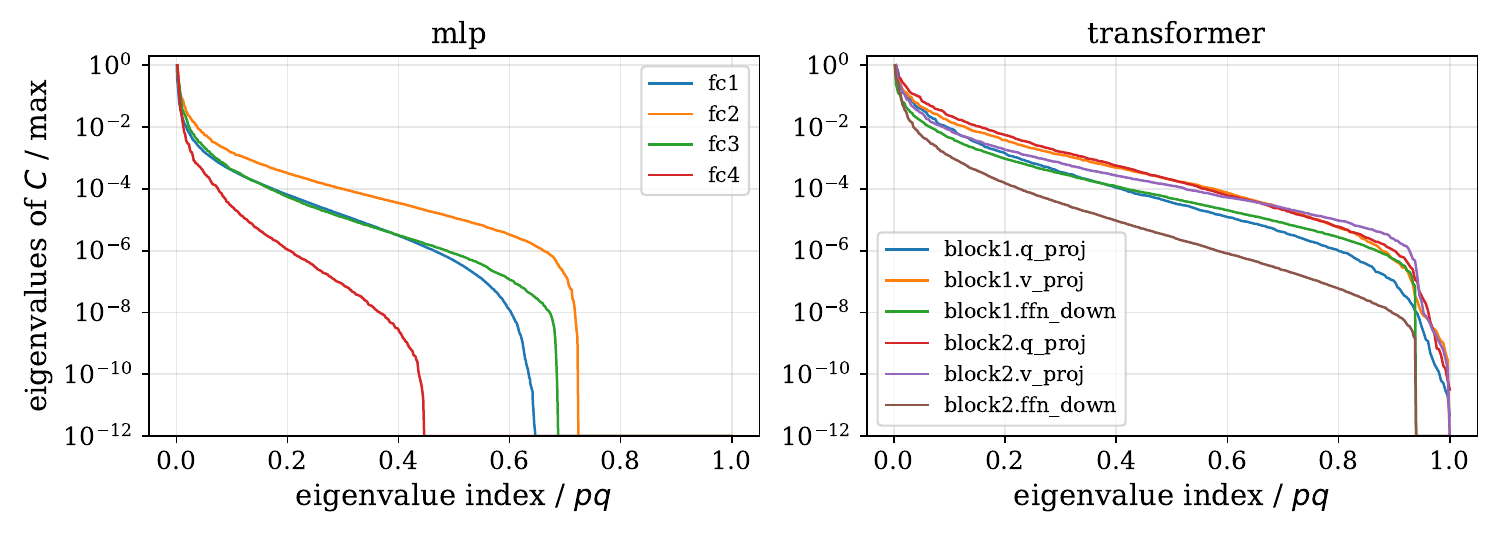}
    \caption{The FB covariance spectra in the rotated-digits task. The left panel shows the FB covariance spectra for the MLP. The right panel shows the FB covariance spectra for the transformer.}
    \label{fig:digit_rot_fb_spectra}
\end{figure}

\begin{figure}[h]
    \centering
    \includegraphics[width=0.85\linewidth]{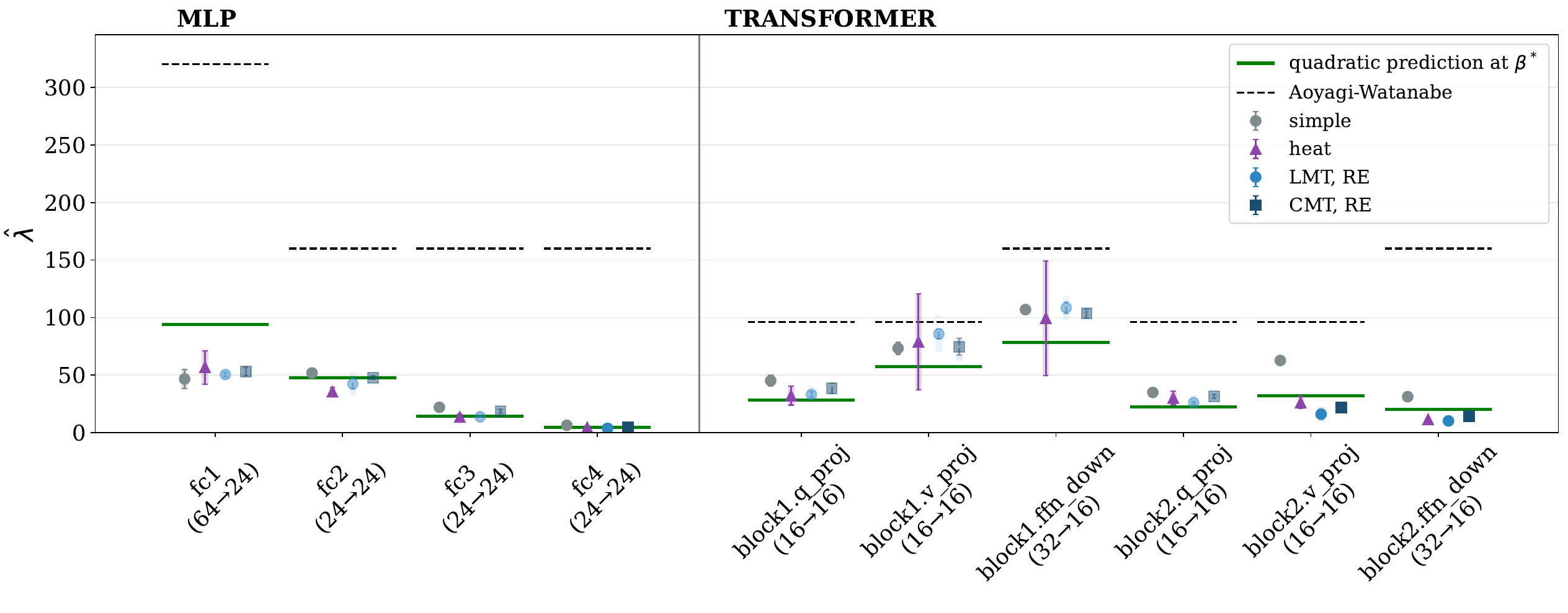}
    \caption{The RLCT estimates based on the localized prior and the MALA in the rotated-digits task. 
    The green dashed lines denote the quadratic
prediction $\lambda_{\mathrm{quad}}(\beta^\star)$ of the Laplace approximation, and the black dashed lines denote the \citet{Aoyagi+Watanabe:2005} theoretical values for the equal rank cases.
    }
    \label{fig:digit_rot_rlct_tl_lora}
\end{figure}

\begin{figure}[h]
    \centering
    \includegraphics[width=0.85\linewidth]{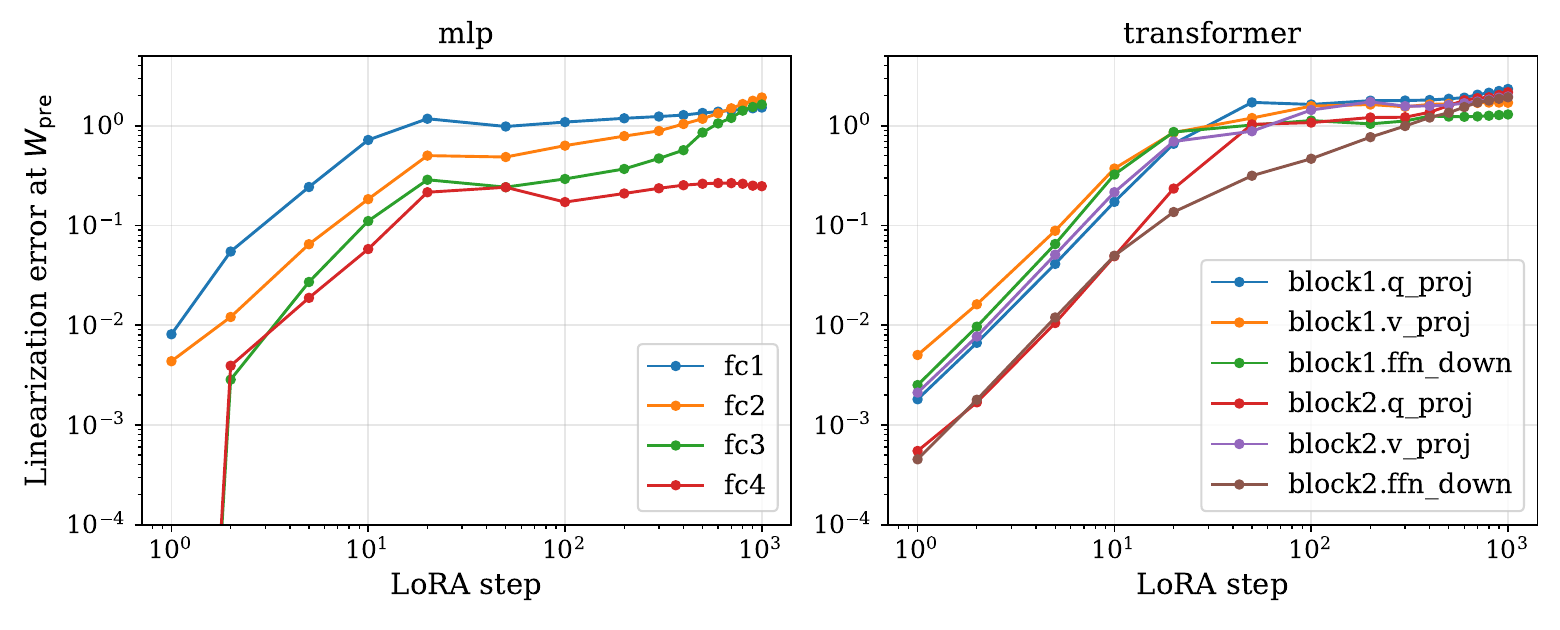}
    \caption{Linearization error in the rotated-digits task. At each LoRA step, the deviation of the change of the model output from the pretrained network from its first-order approximation (the Frobenius inner product between the FB feature at the pretrained weight and the LoRA update),
relative to the norm of that change over the the target-domain training data.}
    \label{fig:digit_rot_linearlizationerror}
\end{figure}

We next consider a transfer task on the $8\times8$ handwritten digits of scikit-learn: a network
pretrained on upright images is adapted to images rotated by $25^\circ$, with the digit value regressed as
$y=(\text{digit}-4.5)/4.5$. We refer to it as the rotated-digits task.
LoRA is fitted on a small target set with $50$ images per digit
($n=500$). 

We consider two architectures.
\begin{itemize}
  \item A four-hidden-layer ReLU MLP with widths $(64,24,24,24,24,1)$: 
  The adapted modules are
  \texttt{fc1}: $\RR^{64}\to\RR^{24}$,
  \texttt{fc2}: $\RR^{24}\to\RR^{24}$,
  \texttt{fc3}: $\RR^{24}\to\RR^{24}$,
  \texttt{fc4}: $\RR^{24}\to\RR^{24}$.
  \item A two-block transformer whose tokens are the $8$ image rows: The input is embedded into dimension $d_{\mathrm{model}}=16$,
  the number of heads is $2$, a feed-forward width is $32$, and a CLS readout is installed. The adapted modules are \texttt{q\_proj} (query projection) and
        \texttt{v\_proj} (value projection) and \texttt{ffn\_down} (output projection of feed-forward network) in each block.
\end{itemize}
LoRA of rank $r=8$ is inserted into one module at a time, all other weights frozen, and fitted by Adam on the squared loss without weight decay.
For the posterior computation, the likelihood is Gaussian with the noise variance set to the training mean squared error at the fitted
   point $\hat w$, and the localized prior $\mathcal N(\hat w,\gamma^{-1}I)$ uses $\gamma=1$.

The FB covariance spectra of the two architectures differ qualitatively (Figure~\ref{fig:digit_rot_fb_spectra}).
In the MLP, the spectrum of every module has an exact kernel: its spectrum drops to numerical zero well before the last index. 
In the transformer, the eigenvalues decay over many orders of magnitude, so the FB covariance $C$ seems to be effectively degenerate.

Figure~\ref{fig:digit_rot_rlct_tl_lora} shows all RLCT estimates per module, together with the quadratic
prediction $\lambda_{\mathrm{quad}}(\beta^\star)$ of the Laplace approximation and the value
$\frac12r(m+d-r)$ of \citet{Aoyagi+Watanabe:2005} with the true rank equal to the LoRA rank.
All estimates lie far below the latter, and most of them agree with $\lambda_{\mathrm{quad}}(\beta^\star)$,
except for \texttt{fc1} and the first transformer block.
They also depend on the layer: within the MLP they decrease from \texttt{fc1} to \texttt{fc4}, and the smallest
value, for \texttt{fc4}, goes with the largest kernel of the FB covariance.

Figure~\ref{fig:digit_rot_linearlizationerror} studies the validity of local linearization in our study.  We see that the relative linearization is small during the early phase of fine-tuning, which supports the use of FB features computed at the pretrained weights as a local description of the adaptation problem.  As training progresses, we see that the error steadily increases and eventually reaches order 1, which indicates that higher-order effects become important as the adapted model moves away from pretrained initialization.  The rate of this diversion depends on layer, since some layers appear to be still approximated well by the linearized model than others.  Thus, our linearized theory should be taken as a characterization of the local singular geometry of LoRAa around the pretrained model, and not so much as an approximation of the full nonlinear fine-tuning trajectory.

\subsection{Real NLP task: Prompt-Based Surgical Fine-Tuning}

\begin{figure}[h]
    \centering
    \includegraphics[width=0.85\linewidth]{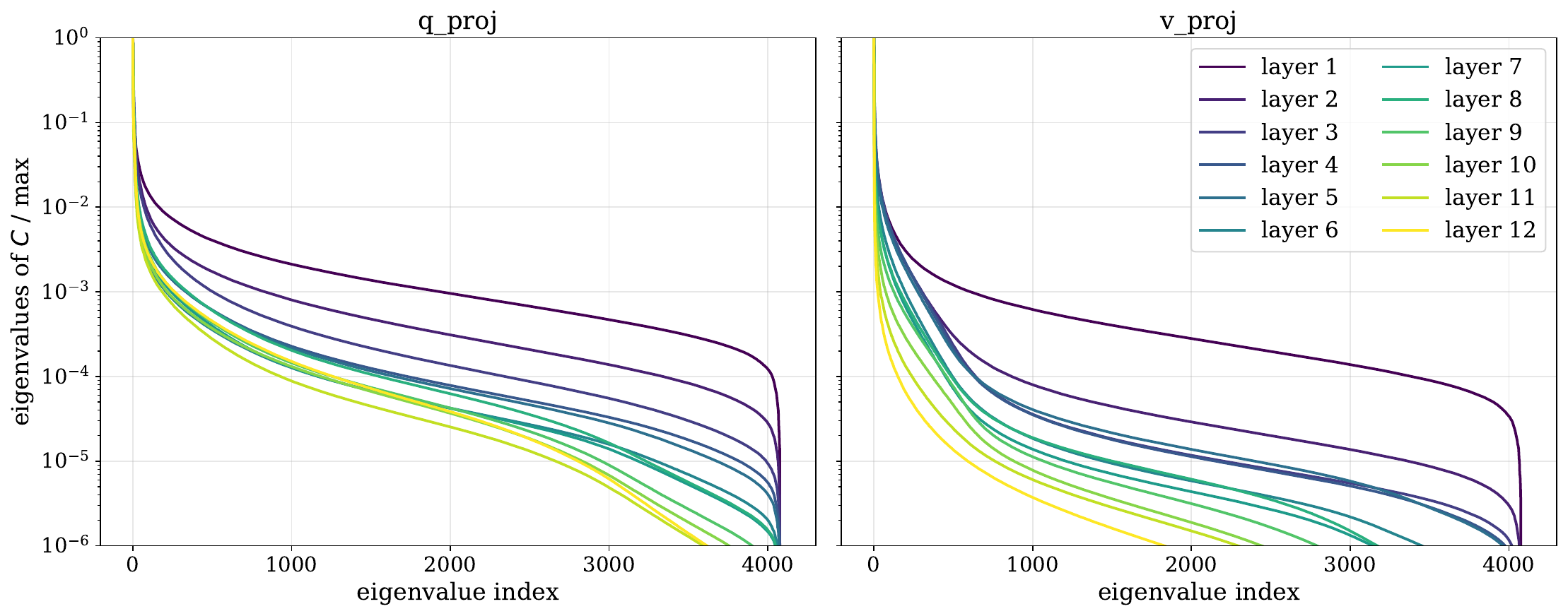}
    \caption{The FB covariance spectra in the prompt-based surgical fine tuning. The left panel shows the FB covariance spectra for the query projection units. The right panel shows the FB covariance spectra for the value projection units. The spectra are computed using 4096 target-domain data.}
    \label{fig:stsb_fb_spectrum}
\end{figure}

\begin{figure}[h]
    \centering
    \includegraphics[width=0.85\linewidth]{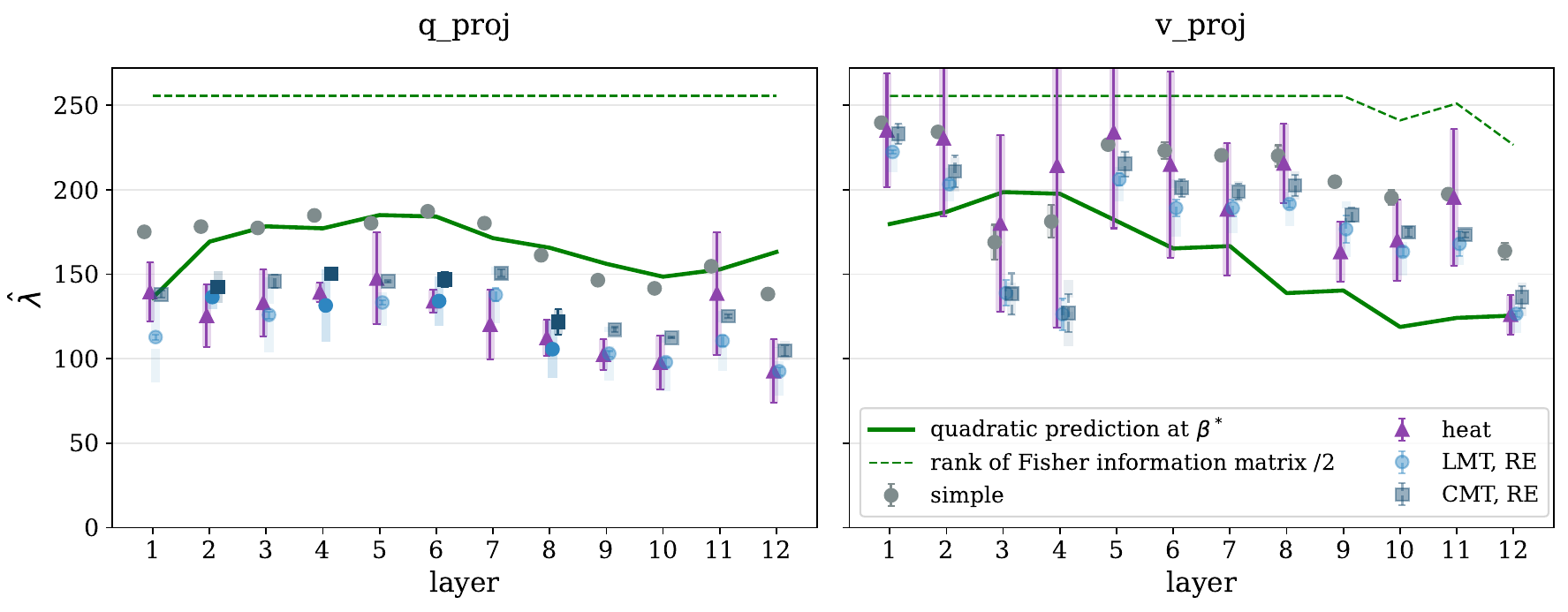}
    \caption{The RLCT estimates based on the localized prior and the MALA in the prompt-based surgical fine tuning. 
    The green solid lines denote the quadratic
prediction $\lambda_{\mathrm{quad}}(\beta^\star)$ of the Laplace approximation, and the green dashed lines denote the ranks of the Fisher information matrices for the LoRA parameters at the fitted LoRA updates.}
    \label{fig:stsb_rlct_per_layer}
\end{figure}

Finally, we conduct an experiment on surgical LoRA fine-tuning \citep[c.f.,][]{lodha2023,ando2026}) to
check the FB covariance spectra and
the behaviors of RLCT estimates layer by layer.
We use the Semantic Textual Similarity Benchmark (STS-B, \cite{cer2017}),
a sentence-pair regression task with similarity scores $Y_{i}\in[0,5]$,
and the RoBERTa base model \citep{liu2020}, an
encoder-only transformer with 12 layers and hidden size 768.

We adopt the prompt-based fine-tuning setting of \cite{gao2021} with a moderate number of shots:
we draw $N=512$ training pairs, 256 below and 256 at or above the median score.
Each sentence pair $X_{i}=(S_{i,1},S_{i,2})$ is concatenated as the prompt
``$S_{i,1}$ \texttt{<mask>} , $S_{i,2}$'',
and the output of the model is the difference of the logits of the label words
``Yes'' and ``No'' at the mask position.
Then the output is fitted to the score mapped to the logit scale,
$T_{i}:=\mathrm{logit}\bigl((Y_{i}+\epsilon)/(5+2\epsilon)\bigr)$ with $\epsilon=1/2$.

Following \cite{jang24d}, we linearize the model at the pretrained weight and
adapt LoRA to a single projection at a time,
under the Gaussian likelihood:
\[
\min_{(A,B)}\sum_{i=1}^{N}\frac{(T_{i}-\theta_{i})^{2}}{2\sigma^{2}}
\quad\text{with}\quad
\theta_{i}:= f_{\mathrm{RoBERTa}}(X_{i}) + s \frobprod{\Phi^{(l,m)}(X_{i}), BA},
\]
where $f_{\mathrm{RoBERTa}}(X_{i})$ is the output at the pretrained weight,
$A\in\RR^{r\times 768}$ and $B\in\RR^{768\times r}$ with $r=8$,
$s=\alpha/r$ with $\alpha=r$, and
$\Phi^{(l,m)}(X_{i}):=\sum_{t} g_{i,t}h_{i,t}^{\transpose}$ is the forward-backward feature
at the $m\in \{\text{query}, \text{value}\}$-projection unit in the $l$-th attention layer,
with $h_{i,t}$ the input of the projection at token $t$ and
$g_{i,t}$ the gradient of $f_{\mathrm{RoBERTa}}(X_{i})$ with respect to its output.
The noise level $\sigma$ is set to the residual standard deviation of
the best affine fit of $T$ by $f_{\mathrm{RoBERTa}}$ on the validation set.

Figure \ref{fig:stsb_fb_spectrum} showcases the FB covariance spectra for query and value units.
The spectrum decays more slowly in shallower layers.
The value units show a clear monotone ordering in depth,
whereas the query units of layers 4--12 have nearly the same spectrum.
Figure \ref{fig:stsb_rlct_per_layer} showcases the RLCT estimates.
All estimates lie below the half of rank of the Fisher information matrix that is bounded by $ N/2=256$, and hence far below the \citet{Aoyagi+Watanabe:2005} value
$\frac{1}{2}r(p+q-r)=6112$ for reduced rank regression.
Also, the RLCT is not determined by the Laplace approximation alone.
Finally, 
both the spectra and the RLCT depend strongly on the layer.

\section{Discussion}
\label{sec:discussion}
In this paper, we studied the learning complexity of LoRA from the perspective of singular learning theory.  Our main conclusion is the complexity is determined by multiple factors and not only the low-rank parameterization by itself.  It turns out to depend on how the geometry of the low-rank update interacts with the statistical geometry from the pretrained network.  Under local linearization, the FB covariance interfaces between these two geometries and determines which perturbations of an adapted weight matrix are statistically observable.  As a result, LoRA modules with the same rank and parameter count can have very different learning complexities depending on where they are located in a network.

Our experimental results suggest that this dependence on the pretrained network is significant in practice.  Both the spectrum of the FB covariance and the RLCT vary significantly across layers and modules.  In particular, degeneracy of the covariance in the ambient weight space does not determine the learning complexity of LoRA alone: what matters is how the degeneracy interacts with the low-rank directions that the adapter sees and interacts with.  More broadly, this distinction may be useful in studying parameter-efficient adaptation, where nominal parameter count by itself gives only a limited description of the effective complexity of the learning problem.

Our work suggests several possible directions for future research.  One such direction is a further in-depth of the FB covariance itself, in the sense that its geometry may depend on model architecture, pretrained representation, data distribution, downstream task, and location of the adapted module.  In our experiments, we observed strong layer dependence which suggests that the FB covariance could provide information or indicate where a model should be adapted.  Selecting layers or modules for fine-tuning also remains an open problem; it would be interesting to study whether the FB geometry or the corresponding learning complexity can help predict which adaptation locations are the most effective and how this may relate to existing criteria for surgical fine-tuning.

Importantly, our theoretical study considers only a single layer at a time.  In practice, LoRA is often applied simultaneously to multiple layers or modules.  Extending our framework to this multiple setting would entail a further study of the associated geometry of each adaptation location, as well as of how the statistically observable directions interact across the network.  More broadly, this raises the question of how learning complexity distributes over components of the pretrained models and how they interact under parameter-efficient fine-tuning.  Our work and proposed framework provides a starting point for such questions.

\section*{Acknowledgments}
K.F.~and A.M.~are supported by the EPSRC AI Hub on Mathematical Foundations of Intelligence: An ``Erlangen Programme'' for AI No. EP/Y028872/1.  K.Y.~is supported by JSPS KAKENHI Grant Numbers
JP24K15120, JP24H00247, JP26K02871, JP26K23255, JP26K00325,
JP26K02869, by JST CREST Grant Number JPMJCR26E3,
by MEXT STAR-E NEXT Project (Japan Grant Number JPJ013735).

\section*{AI Usage Disclosure}
Generative AI was used for the following: supporting language and minor stylistic improvements in the manuscript, assisting with writing and debugging code, proofreading, and identifying missing content from the literature review. All AI outputs were generated under careful human guidance, reviewed and verified by the authors, who take full responsibility for the final content.

All mathematical statements and proofs were generated without the use of any AI. 


\vfill\eject

\bibliographystyle{plainnat}
\bibliography{literature}

@article{jacot2018,
  title={Neural tangent kernel: Convergence and generalization in neural networks},
  author={Jacot, Arthur and Gabriel, Franck and Hongler, Cl{\'e}ment},
  journal={Advances in neural information processing systems},
  volume={31},
  year={2018}
}

@inproceedings{
zhang2025,
title={Lo{RA}-One: One-Step Full Gradient Could Suffice for Fine-Tuning Large Language Models, Provably and Efficiently},
author={Yuanhe Zhang and Fanghui Liu and Yudong Chen},
booktitle={Forty-second International Conference on Machine Learning},
year={2025},
url={https://openreview.net/forum?id=KwIlvmLDLm}
}

@article{dayi2024,
  title={Gradient dynamics for low-rank fine-tuning beyond kernels},
  author={Dayi, Arif and Chen, Sitan},
  journal={arXiv:2411.15385},
  year={2024}
}

@article{watanabe2001,
  title={Algebraic geometrical methods for hierarchical learning machines},
  author={Watanabe, Sumio},
  journal={Neural Networks},
  volume={14},
  number={8},
  pages={1049--1060},
  year={2001},
  publisher={Elsevier}
}

@article{aoyagi2005,
  title={Resolution of singularities and the generalization error with {B}ayesian estimation for layered neural network},
  author={Aoyagi, Miki and Watanabe, Sumio and others},
  journal={IEICE Trans},
  volume={88},
  number={10},
  pages={2112--2124},
  year={2005}
}

@article{aoyagi2012,
  title={Learning coefficient of generalization error in {B}ayesian estimation and {V}andermonde matrix-type singularity},
  author={Aoyagi, Miki and Nagata, Kenji},
  journal={Neural Computation},
  volume={24},
  number={6},
  pages={1569--1610},
  year={2012}
}

@inproceedings{welling2011,
  title={Bayesian learning via stochastic gradient Langevin dynamics},
  author={Welling, Max and Teh, Yee},
  booktitle={Proceedings of the 28th International Conference on Machine Learning},
  pages={681--688},
  year={2011}
}

@article{hukushima1996,
  title={Exchange {M}onte {C}arlo method and application to spin glass simulations},
  author={Hukushima, Koji and Nemoto, Koji},
  journal={Journal of the Physical Society of Japan},
  volume={65},
  number={6},
  pages={1604--1608},
  year={1996},
  publisher={The Physical Society of Japan}
}

@article{sergeant2026,
  title={Exact Algebraic Computation of Learning Coefficients for Two-Dimensional Singular Models},
  author={Sergeant-Perthuis, Gregoire and Tsigaridas, Elias and Tsukahara, Jules},
  journal={arXiv:2608.20183},
  year={2026}
}

@article{takio2025,
  title={Estimation of the Learning Coefficient Using Empirical Loss},
  author={Takio, Tatsuyoshi and Suzuki, Joe},
  journal={arXiv:2502.09998},
  year={2025}
}

@article{hayashi2017,
  title={Upper bound of {B}ayesian generalization error in non-negative matrix factorization},
  author={Hayashi, Naoki and Watanabe, Sumio},
  journal={Neurocomputing},
  volume={266},
  pages={21--28},
  year={2017},
  publisher={Elsevier}
}

@article{aoyagi2024,
  title={Consideration on the learning efficiency of multiple-layered neural networks with linear units},
  author={Aoyagi, Miki},
  journal={Neural Networks},
  volume={172},
  pages={106132},
  year={2024},
  publisher={Elsevier}
}

@article{drton2025,
  title={Singular Learning Theory for Factor Analysis},
  author={Drton, Mathias and Gross, Elizabeth and Kosta, Dimitra and Leykin, Anton and McCormack, Andrew and Sullivant, Seth and Windisch, Daniel},
  journal={arXiv:2511.15419},
  year={2025}
}

@article{watanabe2026,
  title={A Statistical Difference between Single-Layer Learning and Hierarchical Learning in Wide Neural Networks},
  author={Watanabe, Sumio},
  journal={arXiv:2607.23397},
  year={2026}
}

@article{hayashi2025,
  title={Bayesian generalization error in linear neural networks with concept bottleneck structure and multitask formulation},
  author={Hayashi, Naoki and Sawada, Yoshihide},
  journal={Neurocomputing},
  volume={638},
  pages={130165},
  year={2025},
  publisher={Elsevier}
}

@article{kurumadani2026,
  title={Upper Bounds for Local Learning Coefficients of Three-Layer Neural Networks},
  author={Kurumadani, Yuki},
  journal={Neural Networks},
  pages={109374},
  year={2026},
  publisher={Elsevier}
}

@ARTICLE{Drton2017,
  title     = "Marginal likelihood and model selection for {G}aussian latent tree
               and forest models",
  author    = "Drton, Mathias and Lin, Shaowei and Weihs, Luca and Zwiernik,
               Piotr",
  journal   = "Bernoulli",
  volume    =  23,
  number    =  2,
  pages     = "1202--1232",
  year      =  2017
}

@article{yamazaki2003,
  title={Singularities in mixture models and upper bounds of stochastic complexity},
  author={Yamazaki, Keisuke and Watanabe, Sumio},
  journal={Neural networks},
  volume={16},
  number={7},
  pages={1029--1038},
  year={2003},
  publisher={Elsevier}
}

@inproceedings{gao2021,
    title = "Making Pre-trained Language Models Better Few-shot Learners",
    author = "Gao, Tianyu  and
      Fisch, Adam  and
      Chen, Danqi",
    editor = "Zong, Chengqing  and
      Xia, Fei  and
      Li, Wenjie  and
      Navigli, Roberto",
    booktitle = "Proceedings of the 59th Annual Meeting of the {A}ssociation for {C}omputational {L}inguistics and the 11th International Joint Conference on Natural Language Processing (Volume 1: Long Papers)",
    month = aug,
    year = "2021",
    address = "Online",
    publisher = "Association for Computational Linguistics",
    url = "https://aclanthology.org/2021.acl-long.295/",
    doi = "10.18653/v1/2021.acl-long.295",
    pages = "3816--3830"
}

@misc{
liu2020,
title={{RoBERTa: A Robustly Optimized BERT Pretraining Approach}},
author={Yinhan Liu and Myle Ott and Naman Goyal and Jingfei Du and Mandar Joshi and Danqi Chen and Omer Levy and Mike Lewis and Luke Zettlemoyer and Veselin Stoyanov},
year={2020},
url={https://openreview.net/forum?id=SyxS0T4tvS}
}

@book{watanabe2018mathematical,
  title={Mathematical theory of {B}ayesian statistics},
  author={Watanabe, Sumio},
  year={2018},
  publisher={chapman and hall/cRc}
}

@article{imai2019estimating,
  title={Estimating real log canonical thresholds},
  author={Imai, Toru},
  journal={arXiv preprint arXiv:1906.01341},
  year={2019}
}

@article{okuno2023generalization,
  title={A generalization gap estimation for overparameterized models via the {L}angevin functional variance},
  author={Okuno, Akifumi and Yano, Keisuke},
  journal={Journal of Computational and Graphical Statistics},
  volume={32},
  number={4},
  pages={1287--1295},
  year={2023},
  publisher={Taylor \& Francis}
}

@article{hu2021lora,
  title={Lo{R}{A}: {L}ow-rank adaptation of large language models},
  author={Hu, Edward J and Shen, Yelong and Wallis, Phillip and Allen-Zhu, Zeyuan and Li, Yuanzhi and Wang, Shean and Wang, Lu and Chen, Weizhu},
  journal={arXiv:2106.09685},
  year={2021}
}

@article{Aoyagi+Watanabe:2005,
  title = {Stochastic complexities of reduced rank regression in Bayesian estimation},
  volume = {18},
  issn = {0893-6080},
  doi = {10.1016/j.neunet.2005.03.014},
  number = {7},
  journal = {Neural Networks},
  author = {Aoyagi, Miki and Watanabe, Sumio},
  year = {2005},
  month = Sept,
  pages = {924–933},
}

@article {MR3614481,
    AUTHOR = {Lin, Shaowei},
     TITLE = {Ideal-theoretic strategies for asymptotic approximation of
              marginal likelihood integrals},
   JOURNAL = {J. Algebr. Stat.},
  FJOURNAL = {Journal of Algebraic Statistics},
    VOLUME = {8},
      YEAR = {2017},
    NUMBER = {1},
     PAGES = {22--55},
      ISSN = {1309-3452},
   MRCLASS = {62F99 (14E15 62E99)},
  MRNUMBER = {3614481},
MRREVIEWER = {Thomas\ Kahle},
       DOI = {10.18409/jas.v8i1.47},
       URL = {https://doi.org/10.18409/jas.v8i1.47},
}

@article {MR256156,
       IDS = {Atiyah:1970},
    AUTHOR = {Atiyah, Michael},
     TITLE = {Resolution of singularities and division of distributions},
   JOURNAL = {Comm. Pure Appl. Math.},
  FJOURNAL = {Communications on Pure and Applied Mathematics},
    VOLUME = {23},
      YEAR = {1970},
     PAGES = {145--150},
      ISSN = {0010-3640,1097-0312},
   MRCLASS = {46.40 (32.00)},
  MRNUMBER = {256156},
MRREVIEWER = {C.\ P.\ Gupta},
       DOI = {10.1002/cpa.3160230202},
       URL = {https://doi.org/10.1002/cpa.3160230202},
}

@article{Watanabe:2013,
  title = {A widely applicable {B}ayesian information criterion},
  author = {Watanabe, Sumio},
  journal = {Journal of Machine Learning Research},
  volume = {14},
  number = {1},
  pages = {867--897},
  year = {2013},
}

@book{Watanabe:2009,
  title = {Algebraic Geometry and Statistical Learning Theory},
  author = {Watanabe, Sumio},
  series = {Cambridge Monographs on Applied and Computational Mathematics},
  volume = {25},
  publisher = {Cambridge University Press},
  year = {2009},
}

@article{Schwarz:1978,
  title = {Estimating the dimension of a model},
  author = {Schwarz, Gideon},
  journal = {The Annals of Statistics},
  volume = {6},
  number = {2},
  pages = {461--464},
  year = {1978},
}

@article{KiralyTheranTomioka2015,
  author  = {Kir{\'a}ly, Franz and Theran, Louis and Tomioka, Ryota},
  title   = {The Algebraic Combinatorial Approach for Low-Rank Matrix Completion},
  journal = {Journal of Machine Learning Research},
  year    = {2015},
  volume  = {16},
  number  = {41},
  pages   = {1391--1436},
  url     = {http://jmlr.org/papers/v16/kiraly15a.html}
}

@book{MR4321457,
  author    = {Mond, David and Nu\~no-Ballesteros, Juan},
  title     = {Singularities of mappings---the local behaviour of smooth and complex analytic mappings},
  series    = {Grundlehren der mathematischen Wissenschaften},
  volume    = {357},
  publisher = {Springer, Cham},
  year      = {2020},
  pages     = {xv+567},
  isbn      = {978-3-030-34439-9},
  doi       = {10.1007/978-3-030-34440-5},
}

@InProceedings{malladi23a,
  title = 	 {A Kernel-Based View of Language Model Fine-Tuning},
  author =       {Malladi, Sadhika and Wettig, Alexander and Yu, Dingli and Chen, Danqi and Arora, Sanjeev},
  booktitle = 	 {Proceedings of the 40th International Conference on Machine Learning},
  pages = 	 {23610--23641},
  year = 	 {2023},
  editor = 	 {Krause, Andreas and Brunskill, Emma and Cho, Kyunghyun and Engelhardt, Barbara and Sabato, Sivan and Scarlett, Jonathan},
  volume = 	 {202},
  series = 	 {Proceedings of Machine Learning Research},
  publisher =    {PMLR}
}

@inproceedings{cer2017,
  title={SemEval-2017 task 1: Semantic textual similarity multilingual and crosslingual focused evaluation},
  author={Cer, Daniel and Diab, Mona and Agirre, Eneko and Lopez-Gazpio, Inigo and Specia, Lucia},
  booktitle={Proceedings of the 11th international workshop on semantic evaluation (SemEval-2017)},
  pages={1--14},
  year={2017}
}

@InProceedings{jang24d,
  title = 	 {{L}o{RA} Training in the {NTK} Regime has No Spurious Local Minima},
  author =       {Jang, Uijeong and Lee, Jason D. and Ryu, Ernest K.},
  booktitle = 	 {Proceedings of the 41st International Conference on Machine Learning},
  pages = 	 {21306--21328},
  year = 	 {2024},
  editor = 	 {Salakhutdinov, Ruslan and Kolter, Zico and Heller, Katherine and Weller, Adrian and Oliver, Nuria and Scarlett, Jonathan and Berkenkamp, Felix},
  volume = 	 {235},
  series = 	 {Proceedings of Machine Learning Research},
  publisher =    {PMLR}
}

@InProceedings{kim25n,
  title = 	 {{L}o{RA} Training Provably Converges to a Low-Rank Global Minimum Or It Fails Loudly ({B}ut it Probably Won’t Fail)},
  author =       {Kim, Junsu and Kim, Jaeyeon and Ryu, Ernest K.},
  booktitle = 	 {Proceedings of the 42nd International Conference on Machine Learning},
  pages = 	 {30224--30247},
  year = 	 {2025},
  editor = 	 {Singh, Aarti and Fazel, Maryam and Hsu, Daniel and Lacoste-Julien, Simon and Berkenkamp, Felix and Maharaj, Tegan and Wagstaff, Kiri and Zhu, Jerry},
  volume = 	 {267},
  series = 	 {Proceedings of Machine Learning Research},
  publisher =    {PMLR}
}

@inproceedings{zeng2024,
  title={The expressive power of low-rank adaptation},
  author={Zeng, Yuchen and Lee, Kangwook},
  booktitle={International Conference on Learning Representations},
  volume={2024},
  pages={5078--5123},
  year={2024}
}

@inproceedings{
lau2025,
title={{The Local Learning Coefficient: A Singularity-Aware Complexity Measure}},
author={Edmund Lau and Zach Furman and George Wang and Daniel Murfet and Susan Wei},
booktitle={The 28th International Conference on Artificial Intelligence and Statistics},
year={2025},
url={https://openreview.net/forum?id=1av51ZlsuL}
}

@article{wei2022,
  title={Deep learning is singular, and that’s good},
  author={Wei, Susan and Murfet, Daniel and Gong, Mingming and Li, Hui and Gell-Redman, Jesse and Quella, Thomas},
  journal={IEEE Transactions on Neural Networks and Learning Systems},
  volume={34},
  number={12},
  pages={10473--10486},
  year={2022},
  publisher={IEEE}
}

@article{tokuda2022,
  title={Intrinsic regularization effect in {B}ayesian nonlinear regression scaled by observed data},
  author={Tokuda, Satoru and Nagata, Kenji and Okada, Masato},
  journal={Physical Review Research},
  volume={4},
  number={4},
  pages={043165},
  year={2022},
  publisher={APS}
}

@ARTICLE{Roberts1996,
  title     = "Exponential convergence of Langevin distributions and their
               discrete approximations",
  author    = "Roberts, Gareth and Tweedie, Richard",
  journal   = "Bernoulli",
  volume    =  2,
  number    =  4,
  pages     =  341,
  year      =  1996
}

@inproceedings{
lee2023,
title={Surgical Fine-Tuning Improves Adaptation to Distribution Shifts},
author={Yoonho Lee and Annie Chen and Fahim Tajwar and Ananya Kumar and Huaxiu Yao and Percy Liang and Chelsea Finn},
booktitle={The Eleventh International Conference on Learning Representations},
year={2023},
url={https://openreview.net/forum?id=APuPRxjHvZ}
}

@inproceedings{ando2026,
  title={Layer-Wise {LoRA} Fine-Tuning: A Similarity Metric Approach},
  author={Ando Ogawa, Keith and Yamamoto, Bruno Lopes and de Alcantara, Lucas Lauton and Pellicer, Lucas and Costa, Rosimeire Pereira and Bollis, Edson and Costa, Anna Helena Reali and Jordao, Artur},
  booktitle={International Conference on Pattern Recognition},
  pages={370--385},
  year={2026}
}

@inproceedings{lodha2023,
    title = "On Surgical Fine-tuning for Language Encoders",
    author = "Lodha, Abhilasha  and
      Belapurkar, Gayatri  and
      Chalkapurkar, Saloni  and
      Tao, Yuanming  and
      Ghosh, Reshmi  and
      Basu, Samyadeep  and
      Petrov, Dmitrii  and
      Srinivasan, Soundararajan",
    editor = "Bouamor, Houda  and
      Pino, Juan  and
      Bali, Kalika",
    booktitle = "Findings of the {A}ssociation for {C}omputational {L}inguistics: {EMNLP} 2023",
    month = dec,
    year = "2023",
    pages = "3105--3113"
}

@article{swendsen1986,
  title={{Replica Monte Carlo simulation of spin-glasses}},
  author={Swendsen, Robert and Wang, Jian-Sheng},
  journal={Physical review letters},
  volume={57},
  number={21},
  pages={2607},
  year={1986}
}

@inproceedings{wang2025,
  title={Differentiation and specialization of attention heads via the refined local learning coefficient},
  author={Wang, George and Hoogland, Jesse and van Wingerden, Stan and Furman, Zach and Murfet, Daniel},
  booktitle={International Conference on Learning Representations},
  volume={2025},
  pages={101037--101082},
  year={2025}
}

@article{
hoogland2025,
title={Loss Landscape Degeneracy and Stagewise Development in Transformers},
author={Jesse Hoogland and George Wang and Matthew Farrugia-Roberts and Liam Carroll and Susan Wei and Daniel Murfet},
journal={Transactions on Machine Learning Research},
issn={2835-8856},
year={2025},
url={https://openreview.net/forum?id=45qJyBG8Oj}
}

@article{Recht.Fazel.Parrilo:2010,
author = {Recht, Benjamin and Fazel, Maryam and Parrilo, Pablo},
title = {Guaranteed Minimum-Rank Solutions of Linear Matrix Equations via Nuclear Norm Minimization},
journal = {SIAM Review},
volume = {52},
number = {3},
pages = {471-501},
year = {2010},
doi = {10.1137/070697835}
}


\end{document}